\pdfoutput=1

\documentclass[11pt]{article}

\usepackage[final]{acl}

\usepackage{times}
\usepackage{latexsym}

\usepackage[T1]{fontenc}

\usepackage[utf8]{inputenc}

\usepackage{microtype}

\usepackage{inconsolata}

\usepackage{graphicx}

\usepackage[font=footnotesize, skip=3.2pt]{subcaption}
\usepackage{xcolor}
\usepackage[most]{tcolorbox}
\usepackage{multicol}
\usepackage{lipsum}
\usepackage{pifont}
\usepackage{booktabs}
\usepackage{times}
\usepackage[table,x11names]{xcolor}
\usepackage{graphicx}
\usepackage[T1]{fontenc}
\usepackage[utf8]{inputenc}
\usepackage{booktabs}
\usepackage{todonotes}
\usepackage{amsmath,amssymb}
\usepackage{cleveref}
\usepackage{xspace}
\usepackage{soul} %
\usepackage{caption}
\usepackage{multirow,multicol}
\usepackage{ulem}
\usepackage{float}
\usepackage{array}
\usepackage{bm}
\usepackage{xfrac}
\usepackage{tcolorbox} %
\usepackage[shortlabels]{enumitem}
\usepackage[outline]{contour}%
\usepackage{tikz}
\usepackage[table]{xcolor}
\usepackage{pgfplots}
\usepackage{subcaption}
\pgfplotsset{compat=1.18}

\setlist[itemize]{noitemsep,left=0mm}

\definecolor{myorange}{RGB}{249,132,93}
\definecolor{mygreen}{RGB}{85,164,139}

\usepackage{tabularx}

\crefname{lstlisting}{listing}{listings}
\Crefname{lstlisting}{Listing}{Listings}
\crefname{equ}{equation}{equations}
\Crefname{equ}{Equation}{Equations}
\Crefname{algorithm}{Algorithm}{Algorithms}
\crefname{example}{example}{examples}
\Crefname{example}{Example}{Examples}
\crefname{prompt}{prompt}{prompts}
\Crefname{prompt}{Prompt}{Prompts}
\DeclareCaptionType{example}[Example][List of examples]
\DeclareCaptionType{prompt}[Prompt][List of prompts]
\DeclareCaptionType{equ}[Equation][List of equations]

\usepackage{marginnote}

\definecolor{TodoColor}{rgb}{1,0.7,0.6}
\definecolor{TodoColor2}{rgb}{0.7,0.7,0.9}
\definecolor{TodoColor3}{rgb}{0.5,0.8,0.5}

\title{Please Translate Again: Two Simple Experiments on Whether \\Human-Like Reasoning Helps Translation}

 \author{
    {\bf Di Wu\thanks{Equal contribution.}\qquad\quad\qquad}
    {\bf Seth Aycock$^{*}$\qquad\quad}
    {\bf Christof Monz}\\
    Language Technology Lab\\
    University of Amsterdam\\
    \medskip
    \texttt{\{d.wu, s.aycock, c.monz\}@uva.nl}
 }

\begin{document}
\maketitle
\begin{abstract}
Large Language Models (LLMs) demonstrate strong reasoning capabilities for many tasks, often by explicitly decomposing the task via Chain-of-Thought (CoT) reasoning.
Recent work on LLM-based translation designs hand-crafted prompts to decompose translation, or trains models to incorporate intermediate steps.~\textit{Translating Step-by-step}~
\citep{briakou2024translating}, for instance, introduces a multi-step prompt with decomposition and refinement of translation with LLMs, which achieved state-of-the-art results on WMT24 test data.
In this work, we scrutinise this strategy's effectiveness. Empirically, we find no clear evidence that performance gains stem from explicitly decomposing the translation process via CoT, at least for the models on test; and we show prompting LLMs to ``translate again'' and self-refine yields even better results than human-like step-by-step prompting. While the decomposition influences translation behaviour, faithfulness to the decomposition has both positive and negative effects on translation. Our analysis therefore suggests a divergence between the optimal translation strategies for humans and LLMs.

\end{abstract}

\footnotetext[0]{
We release our code and 223k segment and paragraph translations in 8 language pairs from 2 models across 4 steps \href{https://github.com/Sethjsa/COT-MT}{\textbf{here}} to enable further research into the effects of decomposition on translation quality.
}

\section{Introduction}
Large Language Models (LLMs) exhibit strong reasoning capabilities, often characterized by a lengthy, step-by-step decomposition of the question before generating the answer---known as Chain-of-Thought (CoT)~\citep{wei22cot}---along with possible attempts and revisions of the answer, referred to as self-refinement~\citep{madaan2023self,chen2023teaching,pan-etal-2024-automatically}. Both CoT and self-refinement resemble human behaviour when tackling complex problems, e.g. in mathematics.

Driven by recent advancements in LLMs' reasoning capabilities, a trend has developed in improving translation quality through a human-like \textit{decomposition--translation--refinement} paradigm. Here, the source text is decomposed into different aspects including meanings, topics, idiomatic expressions etc., followed by translation drafting and refinement based on these aspects, before generating the final translation.

Some recent work explores pre-translation decomposition, focusing on keywords~\citep{he-etal-2024-exploring} or idioms ~\citep{li2024translate}, aided by external resources.
Others address post-translation refinement, guided by external translation quality assessment~\citep{huang-etal-2024-aligning, ki-carpuat-2024-guiding} or explicit self-evaluation~\citep{feng-etal-2025-tear}.
Refinement can be applied iteratively~\citep{chen-etal-2024-iterative, xu-etal-2024-llmrefine}, and is particularly effective for long document-level translation~\citep{wu2024perhaps}. A key work by~\citet{briakou2024translating} combines pre- and post-translation processes via a fixed 4-step prompting strategy---decomposition (or research), drafting, refinement, and proofreading. Their method shows progressive improvements for long-form translation, achieving state-of-the-art results on WMT24.

\begin{figure*}[t]
    \centering
    \includegraphics[width=1.0\textwidth]{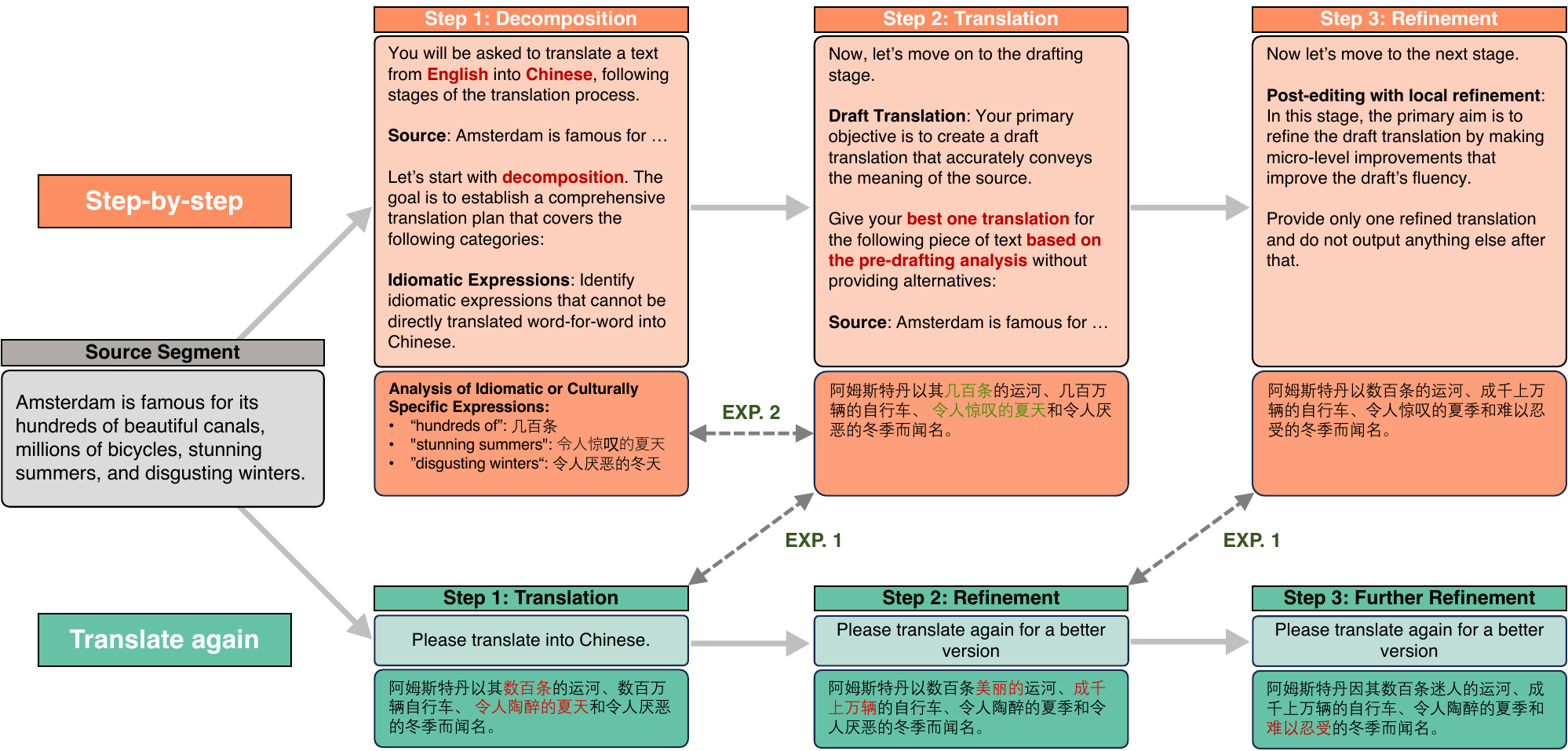}
    \vspace{-1.5em}   
    \caption{Schematic of prompting frameworks for {\color{myorange}\textit{Step-by-step}} translation with decomposition (above) and multi-pass {\color{mygreen}\textit{Translate again}} without decomposition (our method, below), with user prompts and model outputs shown for each step. Experiment 1 (\textsc{exp}. 1) compares translation and refinement outcomes with and without the decomposition step across metrics, input types, and models. Experiment 2 (\textsc{exp}. 2) traces back evidence to assess whether accurately following decomposition improves translation. Full prompts for both settings are provided in Appendix~\ref{ap:prompt_format}.}
    \label{fig:fig-1} 
    \vspace{-1em}        
\end{figure*}

While these studies show performance gains over direct translation in some settings, the generalizability of human-like multi-pass prompting across models and input types remains unclear. Further, most lack an explicit examination or quantitative analysis of the underlying mechanisms behind these gains. To address these points, we design two simple experiments comparing against the current best practice~\citep{briakou2024translating} to answer:
\begin{enumerate}[itemsep=0.2cm, leftmargin=0.5cm]
    \item Does decomposition positively impact translation quality, across models and input types?
    \item How faithful are translations to their decomposition, and does faithfulness improve translations?
\end{enumerate}
\vspace{0.25cm}

Our two experiments find that: (1) most gains come from self-refinement, while decomposition has limited—and sometimes negative—effects, depending largely on the LLM and input type; and (2) decomposition clearly influences translation behaviour, but strict faithfulness to the decomposition does not necessarily improve translation quality.

Given the findings, we encourage the research community to evaluate alternative explanations and reconsider the necessity of human-like decomposition when engaging the reasoning capabilities of LLMs for translation. At a minimum, future studies should consider incorporating a CoT-free refinement strategy---such as the simple `please translate again' prompt used here---as a baseline, given its demonstrated effectiveness and efficiency.

\section{Translation Decomposition and Refinement}
\label{sec:sec-2}
\newcommand{\shiftedincludegraphics}[2]{%
  \makebox[0pt][l]{\hspace*{-#1}\includegraphics[width=\linewidth]{#2}}%
}

\begin{figure*}[t]
    \centering

    \begin{subfigure}[b]{0.23\linewidth}
        \shiftedincludegraphics{0.35em}{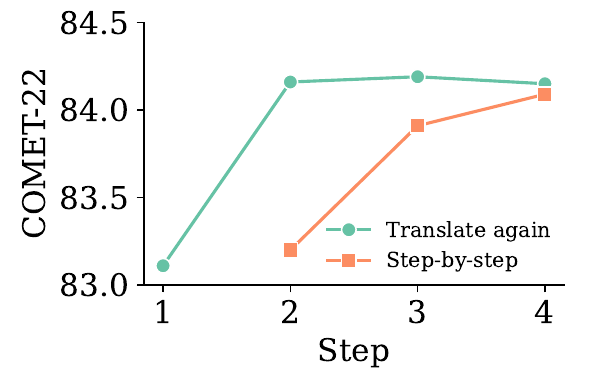}
        \caption{GPT---Segment}
    \end{subfigure}
    \hfill
    \begin{subfigure}[b]{0.23\linewidth}
        \shiftedincludegraphics{0.35em}{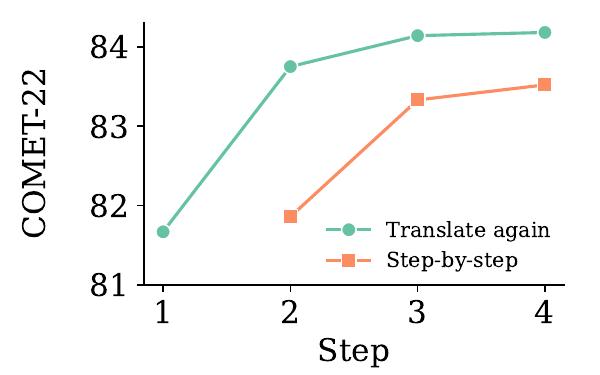}
        \caption{GPT---Paragraph}
    \end{subfigure}
    \hfill
    \begin{subfigure}[b]{0.23\linewidth}
        \shiftedincludegraphics{0.35em}{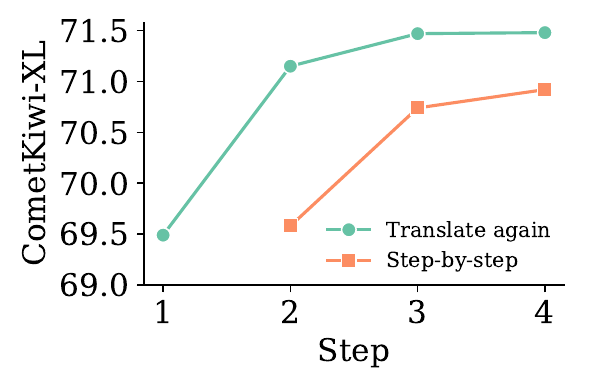}
        \caption{GPT---Segment}
    \end{subfigure}
    \hfill
    \begin{subfigure}[b]{0.23\linewidth}
        \shiftedincludegraphics{0.35em}{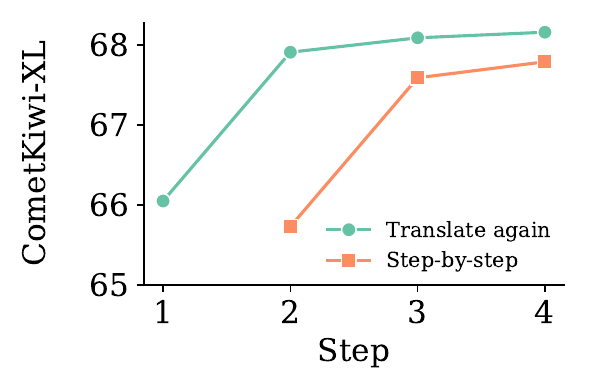}
        \caption{GPT---Paragraph}
    \end{subfigure}

    \vspace{0.75em}

    \begin{subfigure}[b]{0.23\linewidth}
        \shiftedincludegraphics{0.35em}{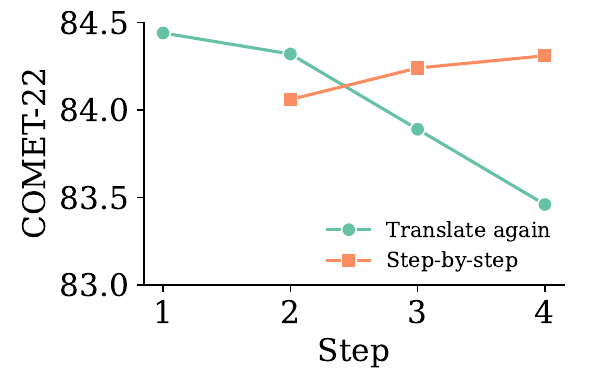}
        \caption{Gemini---Segment}
    \end{subfigure}
    \hfill
    \begin{subfigure}[b]{0.23\linewidth}
        \shiftedincludegraphics{0.35em}{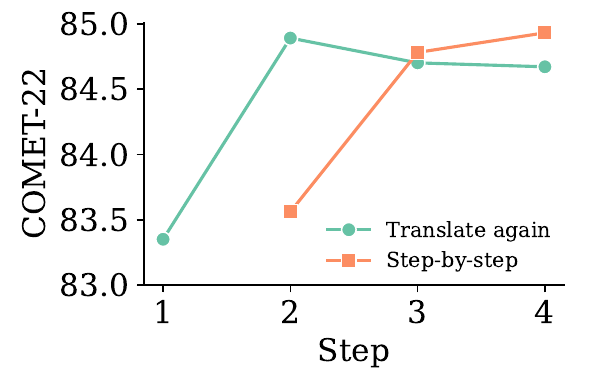}
        \caption{Gemini---Paragraph}
        \label{subfig:gem-doc-com}
    \end{subfigure}
    \hfill
    \begin{subfigure}[b]{0.23\linewidth}
        \shiftedincludegraphics{0.35em}{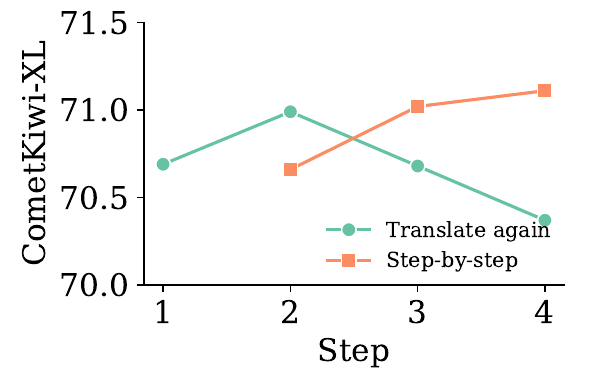}
        \caption{Gemini---Segment}
    \end{subfigure}
    \hfill
    \begin{subfigure}[b]{0.23\linewidth}
        \shiftedincludegraphics{0.35em}{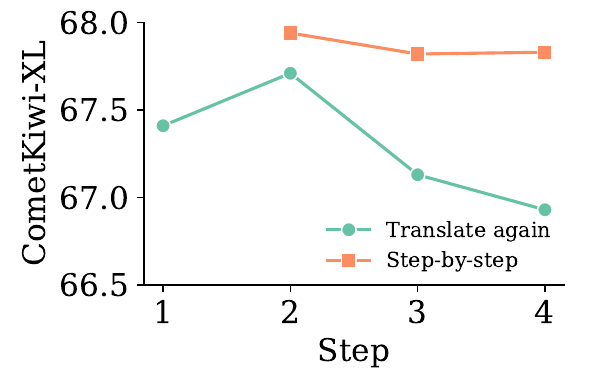}
        \caption{Gemini---Paragraph}
        \label{subfig:gem-doc-kiwi}
    \end{subfigure}

    \caption{{\color{myorange}\textit{Step-by-step}} vs. {\color{mygreen}\textit{Translate again}} results in COMET-22 and CometKiwi-XL for GPT-4o-mini (top) and Gemini-2.0-Flash (bottom), for segment and paragraph-level translation. Note {\color{mygreen}Steps 2--4} iteratively call the LLM to translate again, and {\color{myorange}Step 4} is a proofreading step; see Fig. \ref{fig:fig-1} for an illustration and Appendix \ref{app:full_prompts} for full prompts.}
    \label{fig:comet-kiwi-xl-comparison}
\end{figure*}

Translation by human translators is commonly divided into three phases: pre-drafting, drafting, and post-drafting~\citep{Mossop14-RevisingEditingTranslators}. First, the translator familiarises themselves with the source, consisting of comprehension and planning; next, a full draft translation is written, optionally with the help of external resources; then the translator reviews and revises the draft translation. \citet{briakou2024translating} partially replicate this process with their 4-step prompting process, splitting the final step into refinement and proofreading.

Formally, given a language model \( p_\theta \) and a source text \( x \) to be translated, the output can be viewed as a sample \( O \sim p_\theta(\cdot \mid I(x)) \), where \( I(x) \) is a prompt that may include \( x \) as a component. Multi-step prompting for translation is a sequential process in which the outputs of previous steps are fed into the next prompt. For instance, the \textit{decomposition--translation--refinement} workflow (Figure~\ref{fig:fig-1}, top) can be formalized as:

\vspace{-1.15em}

\begin{align*}
O_d &\sim p_\theta(\cdot \mid I_d(x)), \\
O_t &\sim p_\theta(\cdot \mid I_d(x), O_d, I_t(x)), \\
O_f &\sim p_\theta(\cdot \mid I_d(x), O_d, I_t(x), O_t, I_f(x)).
\end{align*}
\vspace{-1.1em}

\noindent
Here, \( I_d \), \( I_t \), and \( I_f \) denote the prompts for the decomposition, translation, and refinement steps, respectively, and \( O_d \), \( O_t \), and \( O_f \) are their corresponding outputs.  
This study investigates the impact of human-like decomposition (\( O_d \)) on translation quality (\( O_t \)) and final refinement output (\( O_f \)) under varying conditions: (i) model differences (\( \theta \)), (ii) segment- vs.\ paragraph-level source inputs \( x \), and (iii) the presence or absence of \( O_d \) (see Section~\ref{sec:necessity}). We note that while \citeposs{briakou2024translating} fourth step has no access to prior context, this proofreading step produces minimal gains; thus for consistency with our strict self-refinement setting (see Figure \ref{fig:fig-1}), we provide all prior context at each step.
We also explicitly verify whether faithfully following the decomposition generally improves translations (see Section~\ref{sec:verification}).

\section{Experimental Setup}
\paragraph{Models.} We use \verb|GPT-4o-mini|~\citep{OpenAI24-GPT4oMiniAdvancing} and \verb|Gemini-2.0-Flash|~\citep{Google24-IntroducingGemini20}, as performant and cost-effective API LLMs that demonstrate strong reasoning capabilities.

\paragraph{Data.} We use the WMT24++~\citep{deutsch2025wmt24++} dataset as our test set, because a) the dataset was released later than the LLMs we used here, ensuring no data leakage issues, and b) the translation references in WMT24++ are human-written and subsequently post-edited by professional translators, ensuring the highest possible data quality. In this study, we use the post-edited version. We select 8 language pairs (\texttt{en\textrightarrow cs,de,fr,he,ja,ru,uk,zh}) to cover varying writing scripts and families. Each direction shares the same 960 English source samples. For longer-form paragraph-level tests, we combine the segments based on meta-data to give 221 paragraphs (limited to 150 space-separated tokens).

\paragraph{Evaluation.} Following best practice~\citep{kocmi-etal-2024-findings}, we use both reference-based and reference-free neural metrics, using \href{https://huggingface.co/Unbabel/wmt22-comet-da}{COMET$_\mathrm{22}^\mathrm{DA}$} \citep{rei-etal-2022-comet} and \href{https://huggingface.co/Unbabel/wmt23-cometkiwi-da-xl}{COMETKiwi-XL$_\mathrm{23}^\mathrm{DA}$}~\citep{rei-etal-2023-scaling}, respectively. As in the WMT24 Shared Task~\citep{kocmi-etal-2024-findings}, we use the same metrics for paragraph-level evaluation, since they are also effective at this level \citep{deutsch-etal-2023-training}. We also report results in \href{https://huggingface.co/Unbabel/XCOMET-XL}{XCOMET-XL}~\citep{guerreiro2024xcomet} and \href{https://huggingface.co/google/metricx-23-xl-v2p0}{MetricX-23-XL}~\citep{juraska-etal-2023-metricx} in Appendix~\ref{app:full_results_exp1}.

\paragraph{Baselines.} We replicate the \textit{step-by-step} prompt introduced by~\citet{briakou2024translating} as our baseline, and report prompts in full in \Cref{ap:prompt_format}. \citet{briakou2024translating} focus primarily on long-form text using Gemini, whereas we conduct comprehensive experiments on both short- and long-form text and demonstrate generalizability across LLMs.

\paragraph{Proposed method.} We introduce a maximally simple multi-pass prompting method in which the model is asked to produce a translation, then asked to \textit{translate again} and refine 3 more times, given the conversation history, mirroring the step-by-step prompt above. This method involves no explicit pre-drafting step, but expands the number of post-drafting steps arbitrarily, see \Cref{fig:fig-1} (bottom).

\section{Experiment 1: Decomposition's Impact on Translation}
\label{sec:necessity}


\begin{figure*}[t]
    \centering
    \includegraphics[width=0.9\textwidth]{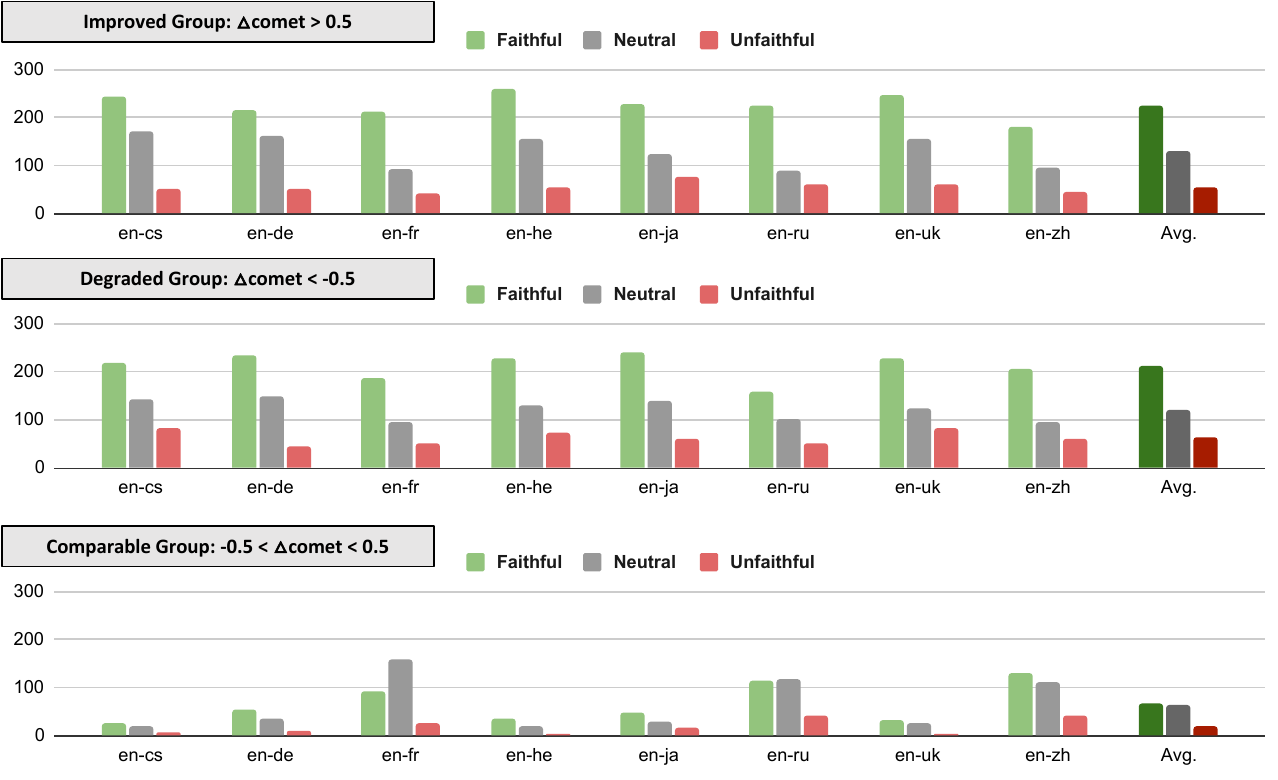}
    \caption{Counts of translations ($t_i^{\textrm{\scriptsize SS}}$) by GPT-4o-mini that are faithful, neutral, or unfaithful to the decomposition, compared to the corresponding direct translation ($t_i^{\textrm{\scriptsize TA}}$). \textit{Avg.} denotes the average over all 8 language directions.}
    \label{fig:fig-win-rate} 
    \vspace{-1.0em}        
\end{figure*}

\definecolor{myorange}{RGB}{249,132,93}
\definecolor{mygreen}{RGB}{85,164,139}

We investigate the effect of decomposition on translation by testing the baseline method ({\color{myorange}\textit{Step-by-step}}) \textbf{with decomposition} against our simple multi-pass prompting method ({\color{mygreen}\textit{Translate again}}) \textbf{without decomposition}. 
\Cref{fig:comet-kiwi-xl-comparison} presents mean step-wise results; see App. Figure~\ref{fig:radar-model-all} for detailed results across languages. Our findings are as follows:

\paragraph{Decomposition.} Comparing {\color{myorange}Step 2} (with decomposition) against {\color{mygreen}Step 1} (without decomposition) shows, at best, a marginally positive effect at the paragraph-level, particularly with Gemini (cf. (f), (h)). This suggests decomposition is not a generally effective strategy for LLM-based translation.

\paragraph{Self-refinement.} Results after a single step of self-refinement show that simply prompting the model to \textit{translate again} for a better version ({\color{mygreen}Step 2}) \textit{without} decomposition consistently yields improvements for GPT-4o-mini over Step-by-step prompting \textit{with} a pre-drafting step ({\color{myorange}Step 3}).

\paragraph{Successive refinement.} Additional steps of refinement, {\color{mygreen}Steps 3--4}, produce only marginal improvements, or occasional degradation for Gemini. We attribute this to the strong performance achieved after 1 refinement step, which may already maximise the LLMs' parametric capabilities and therefore leaves little space for further gains. This also suggests allowing early exit from self-refinement may improve overall performance.

\paragraph{Segment vs. Paragraph-level.} Our findings hold consistently across both segment- and paragraph-level translation. Moreover, we observe that refinements yield slightly larger score improvements for longer-form translation compared to the segment level, in line with \citet{briakou2024translating}.

We therefore find no compelling evidence in favour of human-like, CoT decomposition for translation. Instead, we observe that directly prompting LLMs yields leading results at both the segment and paragraph levels, with the best results achieved after a single step of self-refinement.

\section{Experiment 2: Attribution Analysis of Decomposition}
\label{sec:verification}

We explicitly verify via an attribution analysis whether the decomposition step substantially influences translation behaviour in the subsequent step. We also analyse whether faithfulness to the decomposition results in improved translations.

\paragraph{Explicit verification.} Formally, for a source sentence \(s_i\), we construct a four-tuple \((s_i, d_i, t_i^{\textrm{\scriptsize SS}}, t_i^{\textrm{\scriptsize TA}})\) by prompting an LLM (1) with decomposition \(d_i\), resulting in \(t_i^{\textrm{\scriptsize SS}}\) ({\color{myorange}Step 2}), and (2) without decomposition, resulting in \(t_i^{\textrm{\scriptsize TA}}\) ({\color{mygreen}Step 1}). Explicit verification with an LLM-as-a-judge proceeds as follows:

\begin{itemize}
    \item \textbf{Differentiation:} LLM annotators are asked to identify the main pairs of differences, $\{\{v_1^{\textrm{\scriptsize SS}},v_1^{\textrm{\scriptsize TA}}\}, \{v_2^{\textrm{\scriptsize SS}},v_2^{\textrm{\scriptsize TA}}\}, ..., \{v_k^{\textrm{\scriptsize SS}},v_k^{\textrm{\scriptsize TA}}\}\}$, between translations $t_i^{\textrm{\scriptsize SS}}$ and $t_i^{\textrm{\scriptsize TA}}$.
    \item \textbf{Attribution:} LLM annotators are asked, for each element in the pair of differences $v_i$ between $t_i^{\textrm{\scriptsize SS}}$ and $t_i^{\textrm{\scriptsize TA}}$, how many can be attributed to the decomposition step $d_i$, giving trace-back counts $c_i^{\textrm{\scriptsize SS}}$ and $c_i^{\textrm{\scriptsize TA}}$ respectively; n.b. $t_i^{\textrm{\scriptsize TA}}$ is generated without $d_i$ meaning any attributed differences are coincidental, thus this serves as a baseline.
    \item \textbf{Assessment:} We measure the influence of decomposition \( d_i \) on translation \( t_i^{\textrm{\scriptsize SS}} \) by comparing \( c_i^{\textrm{\scriptsize SS}} \) and \( c_i^{\textrm{\scriptsize TA}} \), where \( c_i^{\textrm{\scriptsize SS}} > c_i^{\textrm{\scriptsize TA}} \) indicates a translation which is \emph{faithful} to the decomposition; \( c_i^{\textrm{\scriptsize SS}} = c_i^{\textrm{\scriptsize TA}} \) indicates a \emph{neutral} translation which is neither faithful nor unfaithful; and \( c_i^{\textrm{\scriptsize SS}} < c_i^{\textrm{\scriptsize TA}} \) indicates an \emph{unfaithful} translation.

\end{itemize}

\noindent We categorise all WMT24-derived four-tuples into \textit{Improved}, \textit{Comparable}, and \textit{Degraded Translation} groups based on the COMET scores of \(t_i^{\textrm{\scriptsize SS}}\) vs. \(t_i^{\textrm{\scriptsize TA}}\). For each group, we conduct explicit verification using \texttt{GPT-4o} as a judge (see Appendix~\ref{app-sec:llm-as-a-judge} for details).
Note that both translations ($t_i^{\textrm{\scriptsize SS}}$ and $t_i^{\textrm{\scriptsize TA}}$) under evaluation are generated by GPT-4o-mini, so no bias from the judge toward either text is expected.
~\Cref{fig:fig-win-rate} shows verification results across groups and directions. We find that:

\paragraph{Translation is mostly faithful to decomposition.} Across all categories and languages, translations conditioned on decompositions contain substantially more differences that can be clearly attributed to the decomposition context (Faithful vs. Unfaithful), compared to direct translations. This suggests that in most cases, translations follow the decomposition produced by the model.

\paragraph{Faithfulness does not improve translation.} 
The group of degraded translations shows a comparable proportion of segments which are influenced by the context compared to the proportion within the improved group of translations.
It suggests the performance impact of decomposition is not stable, and the overall effect is neutral.

Our analysis shows that while decomposition consistently influences translation behaviour, the \textit{positive} impact of decomposition on translation is minimal. 
We tentatively attribute this to the fact that, alongside useful information, the decomposition step may contain errors, which can propagate to the downstream translation task.

\section{Discussion}

Our results suggest that, unlike symbolic tasks such as programming and mathematical reasoning \citep{Sprague25-CoTNotCoT}, translation benefits weakly, if at all, from CoT prompting. Intuitively, in symbolic tasks such as mathematical reasoning, generating intermediate steps with CoT helps the model address compositional reasoning problems to support the final answer. In contrast, translation relies more on holistic language understanding and fluency.

Recent translation work has introduced decomposition with a primary motivation of handling `difficult' lexical choices such as non-compositional idioms. However, we see no intuitive advantage in pre-selecting lexical options in context over direct generation, given that we use the same model for both steps; this is backed up by our empirical observations.

It may be the case that for document-level translation spanning multiple paragraphs, explicit lexical suggestions prior to translation could improve the overall consistency of the output across paragraphs, such as entity names or terminologies. Our study suggests this is likely, since translations are mostly faithful to the decomposition.
However, this exploration lies beyond the scope of the present study and we leave it to future work.

Finally, `reasoning' in the context of LLMs lacks a single clear definition. Current training of reasoning models~\citep{muennighoff2025s1,guo2025deepseek} typically involves multiple components, such as chain-of-thought, reflection, and test-time scaling, and is often coupled with reinforcement learning to promote (objective) alignments. We suggest that future work on reasoning for translation should carefully examine and disentangle the effectiveness of each component.

\section{Conclusion}

We find that CoT reasoning with a decomposition step does not help translation as much as simple self-refinement.
Our results suggest a divergence between the optimal translation strategies for humans and LLMs: while human translators benefit from decomposing the task, LLMs see no clear benefit from CoT reasoning for translation, and imposing human biases may lead to suboptimal outcomes. Further, faithfulness to the generated decomposition does not always yield positive effects. In fact, our maximally simple setting of direct translation and self-refinement (\textit{translate again}) achieves performance comparable to, or even exceeding, the state-of-the-art multi-pass prompting method~\citep{briakou2024translating} at both segment and paragraph levels.
This corroborates findings from the related task of translation from a grammar book~\citep{Aycock24-CanLLMsReallya} that for translation, LLMs exhibit different reasoning tendencies to humans.

\section{Limitations}
We note the following limitations of this work: Due to constrained resources, our investigation primarily focuses on two state-of-the-art LLM families: GPT-4o and Gemini. For Experiment 2, while we observe that GPT-4o is a competent judge in our explicit verification experiments, we note that incorporating judgments from different model families would strengthen the reliability of our results. 

\section{Acknowledgements}
This work was funded in part by the UvA’s Language Sciences for Social Good project, the City of
Amsterdam, and the Netherlands Organization for Scientific Research (NWO) under project numbers
VI.C.192.080 and 2023.017.
We thank our colleagues at the University of Amsterdam, especially Yibin Lei and Xin Sun, for their insightful discussion. D.W. thanks Chongyang for its invaluable spiritual support.
The authors thank the anonymous reviewers for their constructive efforts to improve this research.

\bibliography{custom}

\clearpage
\appendix

\section{Full Results for Experiment 1}
\label{app:full_results_exp1}

In this section, we provide all supplementary results for Experiment 1 (Section~\ref{sec:necessity}). 

Tables~\ref{tab:seg_comet}--\ref{tab:doc_xcomet-xl} show translation results across languages at the segment and paragraph-level, for COMET-22, CometKiwi-23-XL, MetricX-23-XL, and XCOMET-XL.

Figure~\ref{fig:metricx-xcomet-xl-comparison} presents the mean results across languages for GPT-4o-mini and Gemini-2.0-Flash in MetricX and XCOMET-XL under both \textit{step-by-step} and \textit{translate again} prompting strategies.

Figure~\ref{fig:radar-model-all} presents the results of zero-shot (direct) translation and subsequent refinement under both the \textit{Step-by-step} ({\color{myorange}Step 3}) and \textit{Translate again} ({\color{mygreen}Step 2}) strategies on GPT-4o-mini and Gemini-2.0-Flash. We observe across languages and metrics that: 
1) Refinement consistently improves performance over direct translation for both strategies;  
2) The \textit{translate again} strategy generally outperforms the \textit{step-by-step} strategy.

\begin{figure*}[t]
    \centering
    \hspace*{0.4cm} 
    \begin{subfigure}[t]{0.48\linewidth}
        \centering
        \includegraphics[width=\linewidth]{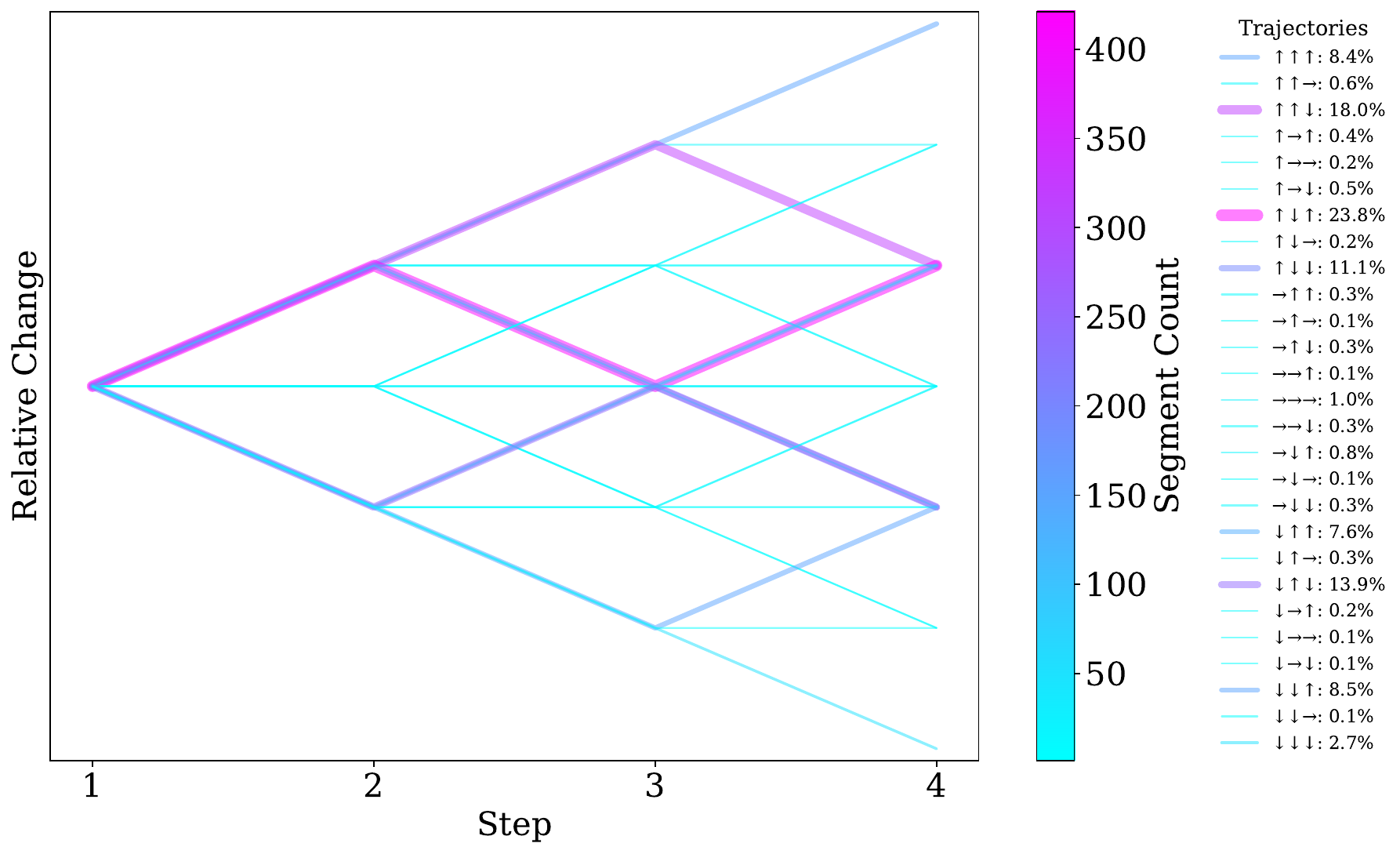}
        \label{fig:traj-gpt-mp-seg}
    \end{subfigure}
    \hfill
    \begin{subfigure}[t]{0.48\linewidth}
        \centering
        \includegraphics[width=\linewidth]{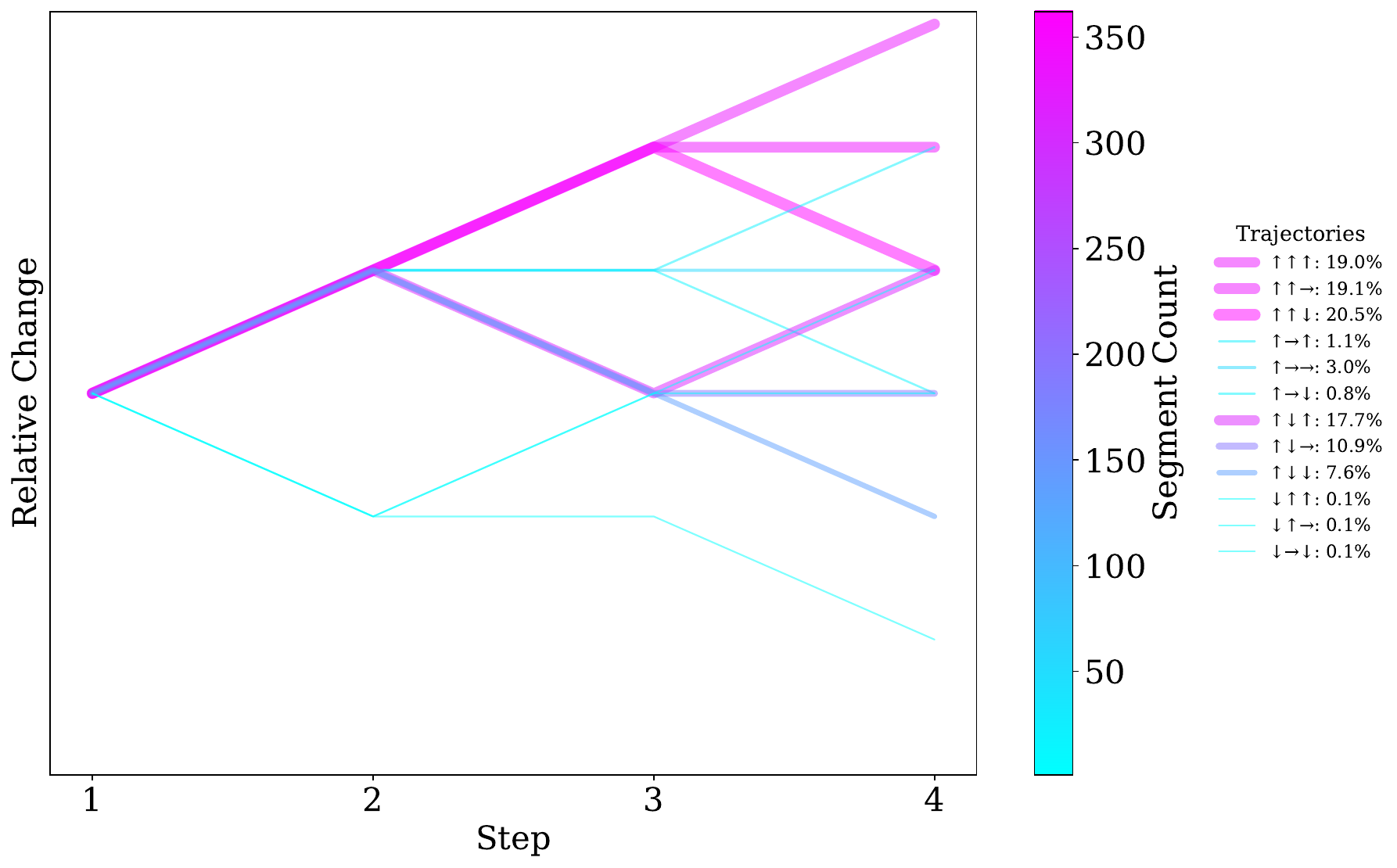}
        \label{fig:traj-gem-seg}
    \end{subfigure}
    \caption{Segment-level COMET score trajectories for GPT-4o-mini with Translate again (left) and Step-by-step (right) prompting strategies. An increase or decrease in the y-axis indicates a \textit{relative} COMET score improvement or degradation compared to the previous step, respectively. Trajectory proportions are shown in the legend.}
    \label{fig:traj-edit-combined-seg}
\end{figure*}

\begin{figure*}[t]
    \centering
    \hspace*{0.4cm}
    \begin{subfigure}[t]{0.48\linewidth}
        \centering
        \includegraphics[width=\linewidth]{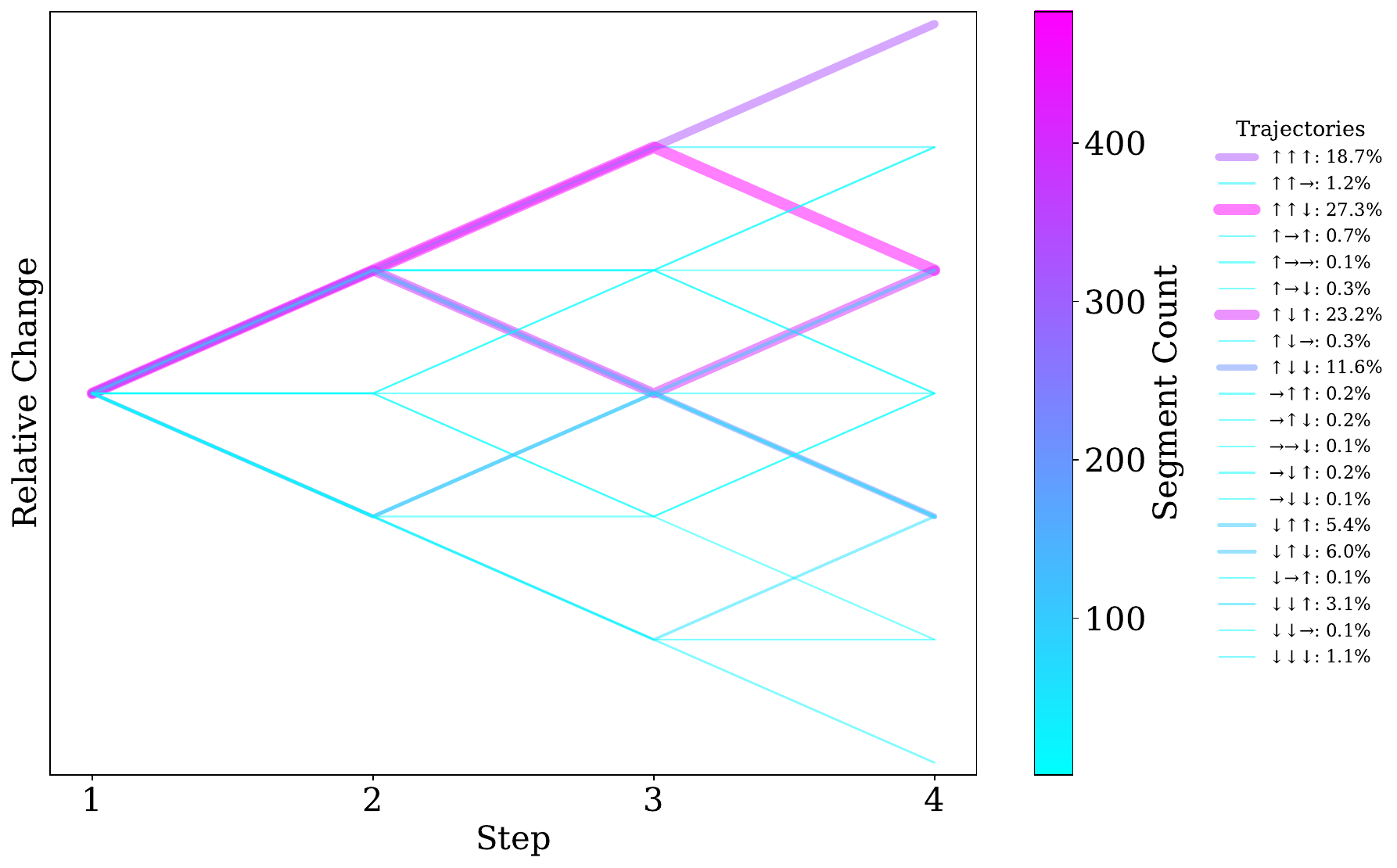}
        \label{fig:traj-gpt-doc}
    \end{subfigure}
    \hfill
    \begin{subfigure}[t]{0.48\linewidth}
        \centering
        \includegraphics[width=\linewidth]{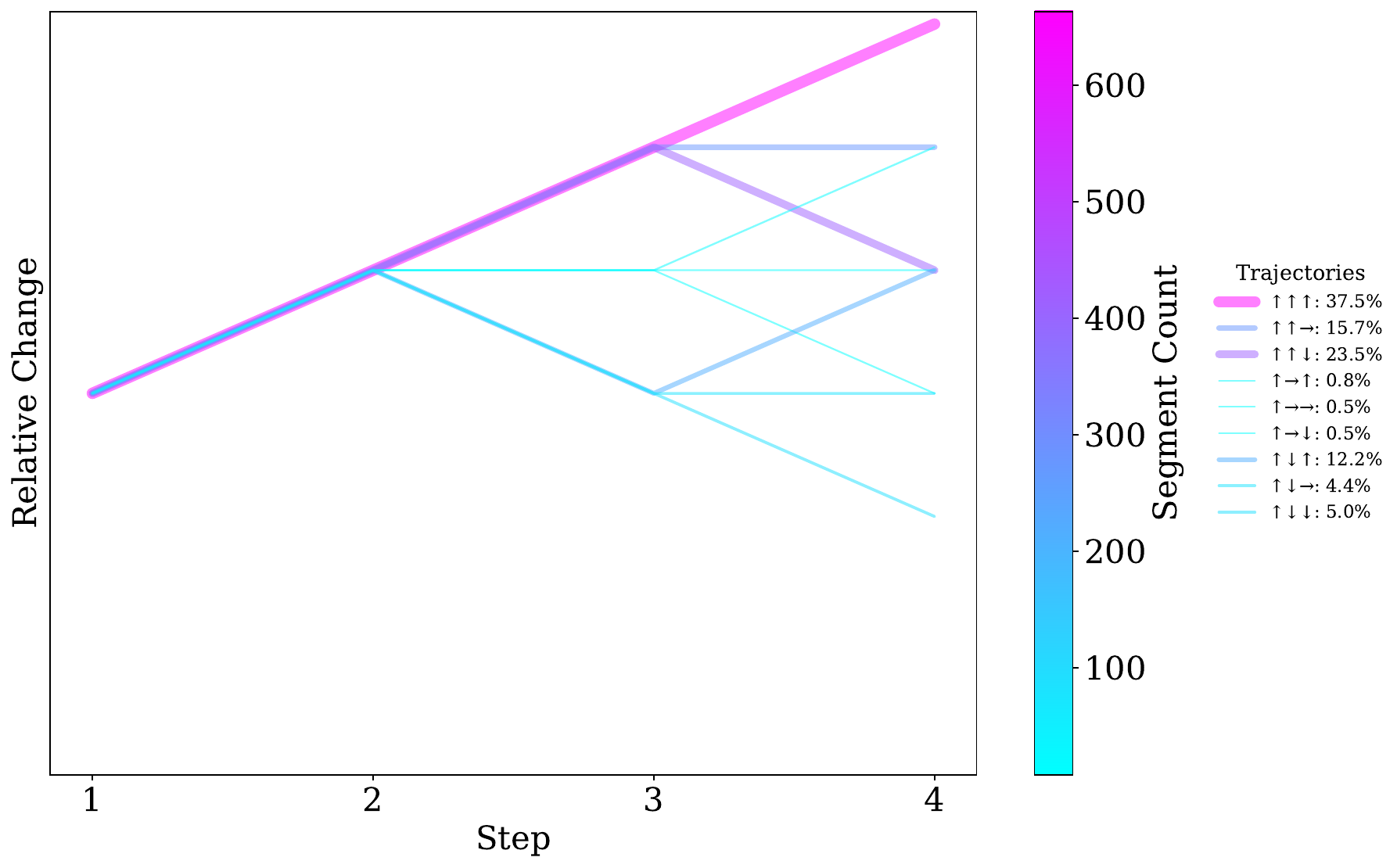}
        \label{fig:traj-gem-doc}
    \end{subfigure}
    \caption{Paragraph-level COMET score trajectories for GPT-4o-mini with Translate again (left) and Step-by-step (right) prompting strategies. An increase or decrease in the y-axis indicates a \textit{relative} COMET score improvement or degradation compared to the previous step, respectively. Trajectory proportions are shown in the legend.}
    \label{fig:traj-edit-combined-doc}
\end{figure*}

Figures \ref{fig:traj-edit-combined-seg} and \ref{fig:traj-edit-combined-doc} show COMET score trajectories for GPT-4o-mini at the segment- and paragraph-level respectively. An increase in the y-axis represents a \textit{relative} increase in COMET score compared to the \textit{previous} step, while a downwards trajectory indicates a \textit{relative} decrease in COMET score. We observe that \textit{translate again} prompting increases many scores from step 1--2, and many paragraphs benefit further from step 2--3. \textit{Step-by-step} shows most segments and paragraphs improve from step 2--3; n.b. for \textit{Step-by-step} we discount steps 1--2 as no translation is produced at step 1. At the segment level, trajectories from step 3--4 are somewhat equally split, while at the paragraph-level most trajectories see further relative improvements.

\section{All Prompt Templates}
\label{app:full_prompts}
\label{ap:prompt_format}

\subsection{\textit{Step-by-Step} Prompts}
The templates for \textit{step-by-step} prompting comprise the Decomposition stage (Figure~\ref{app:step-by-step-1}), the Translation stage (Figure~\ref{app:step-by-step-2}), the Refinement stage (Figure~\ref{app:step-by-step-3}), and the Proofreading stage (Figure~\ref{app:step-by-step-4}).

\subsection{\textit{Translate Again} Prompts}
The templates for \textit{translate again} prompting include the Translation stage (Figure~\ref{app:translate-again-1}) and the Refinement stage (Figure~\ref{app:translate-again-2}). The refinement prompt can be applied iteratively within a session to perform multiple steps of refinement.

\subsection{\textit{LLM-as-a-Judge} Prompts}
\label{app-sec:llm-as-a-judge}
\begin{figure*}[t]
    \centering
        \includegraphics[width=0.8\textwidth]{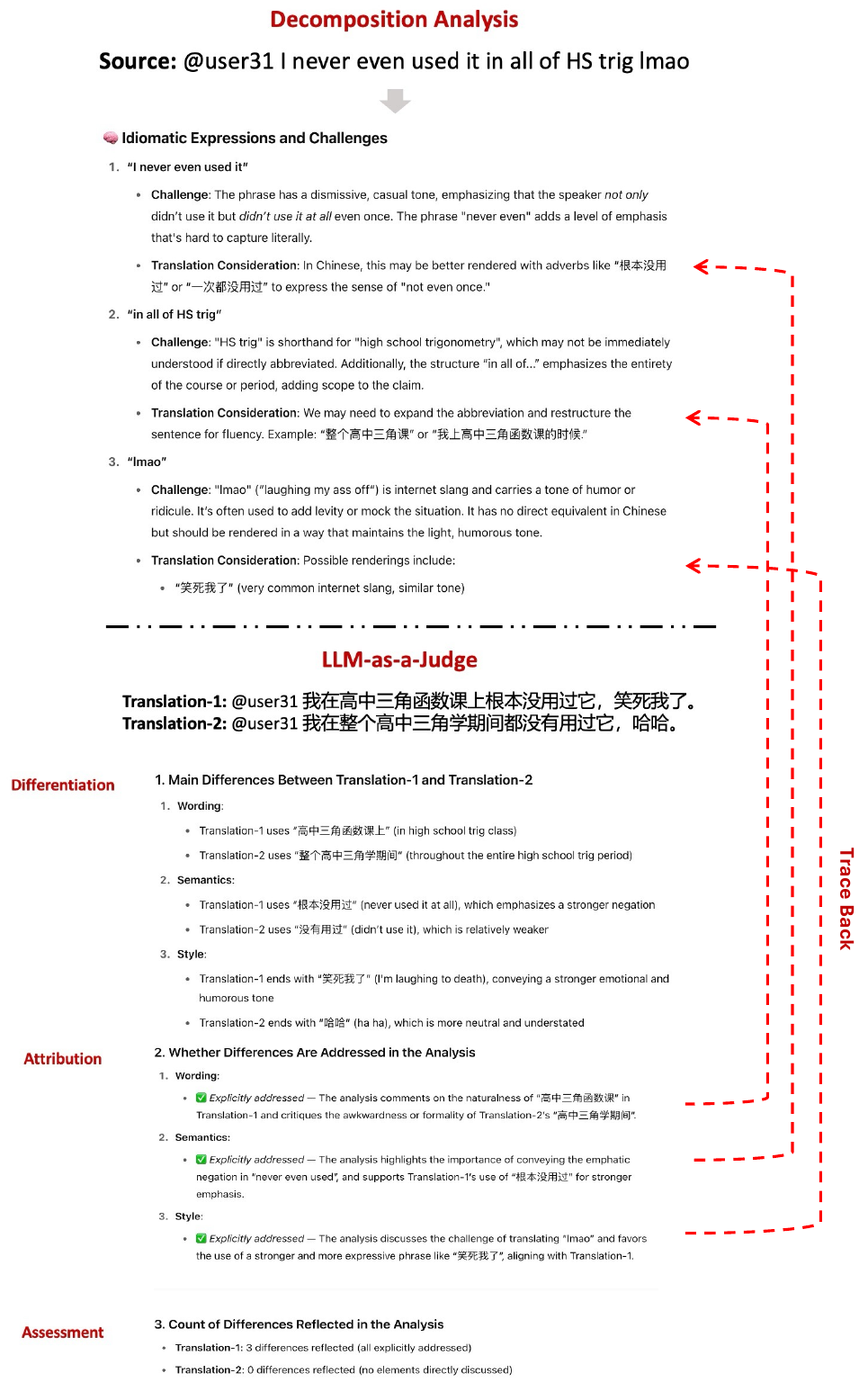}
    \caption{An illustration of using LLM-as-a-Judge to explicitly assess the impact of decomposition on translation behaviour. Given a source text and its corresponding decomposition (analysis results), GPT-4o is employed for three tasks: (1) \textbf{Differentiation} — identifying the differences between Translation 1 and Translation 2; (2) \textbf{Attribution} — mapping each translation difference back to specific elements of the decomposition; and (3) \textbf{Assessment} — evaluating the influence of the decomposition by measuring how many of the differences can be traced back to it.}
    \label{fig:fig-llm-as-a-judge} 
    \vspace{-1.5em}        
\end{figure*}

Figure~\ref{app:llm-as-a-judge} provides the prompt used for LLM-as-a-Judge in Experiment 2 (Section~\ref{sec:verification}). 
We also showcase the output of our \textit{LLM-as-a-judge} in Figure~\ref{fig:fig-llm-as-a-judge}, illustrating how it operates.

\clearpage

    \definecolor{myorange}{RGB}{249,132,93}
    \definecolor{mygreen}{RGB}{85,164,139}
    \begin{figure*}[t]
    \centering

    \begin{subfigure}[b]{0.23\linewidth}
        \includegraphics[width=\linewidth]{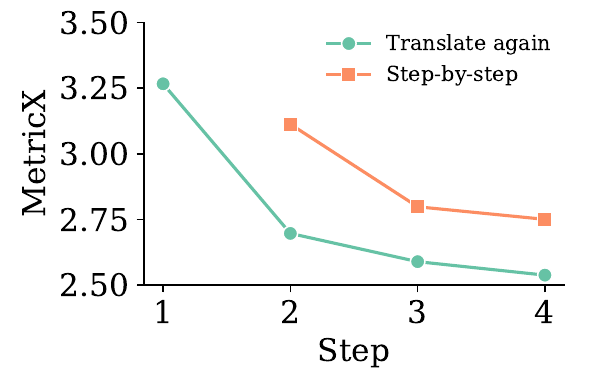}
        \caption{GPT---Segment}
    \end{subfigure}
    \hfill
    \begin{subfigure}[b]{0.23\linewidth}
        \includegraphics[width=\linewidth]{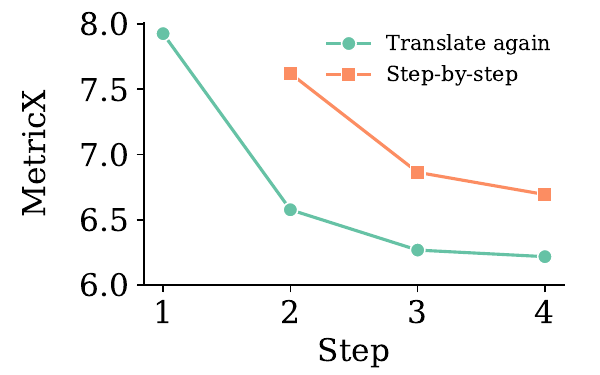}
        \caption{GPT---Paragraph}
    \end{subfigure}
    \hfill
    \begin{subfigure}[b]{0.23\linewidth}
        \includegraphics[width=\linewidth]{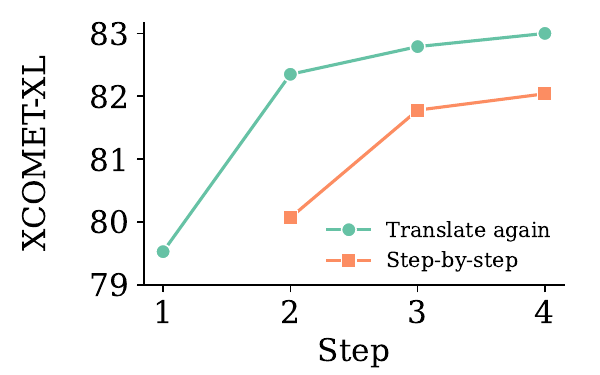}
        \caption{GPT---Segment}
    \end{subfigure}
    \hfill
    \begin{subfigure}[b]{0.23\linewidth}
        \includegraphics[width=\linewidth]{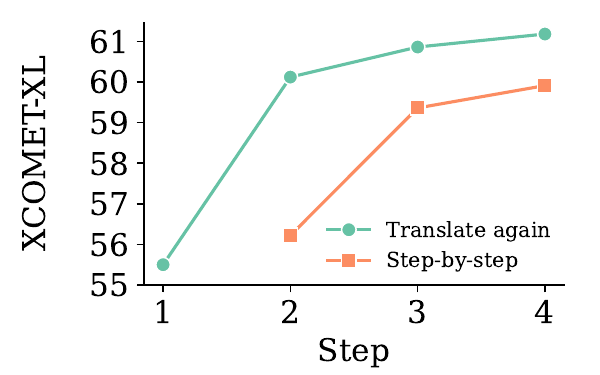}
        \caption{GPT---Paragraph}
    \end{subfigure}

    \vspace{1em}

    \begin{subfigure}[b]{0.23\linewidth}
        \includegraphics[width=\linewidth]{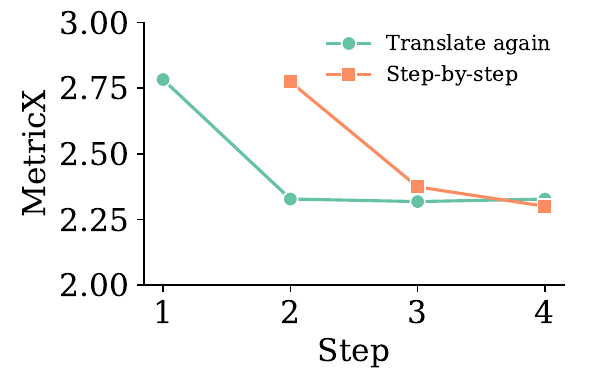}
        \caption{Gemini---Segment}
    \end{subfigure}
    \hfill
    \begin{subfigure}[b]{0.23\linewidth}
        \includegraphics[width=\linewidth]{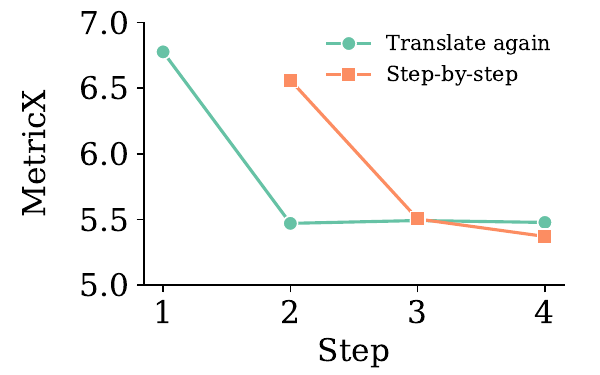}
        \caption{Gemini---Paragraph}
    \end{subfigure}
    \hfill
    \begin{subfigure}[b]{0.23\linewidth}
        \includegraphics[width=\linewidth]{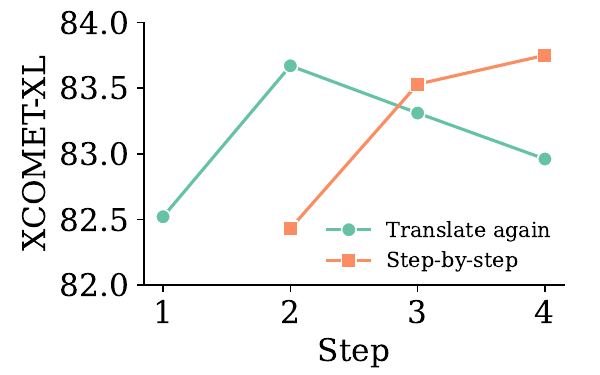}
        \caption{Gemini---Segment}
    \end{subfigure}
    \hfill
    \begin{subfigure}[b]{0.23\linewidth}
        \includegraphics[width=\linewidth]{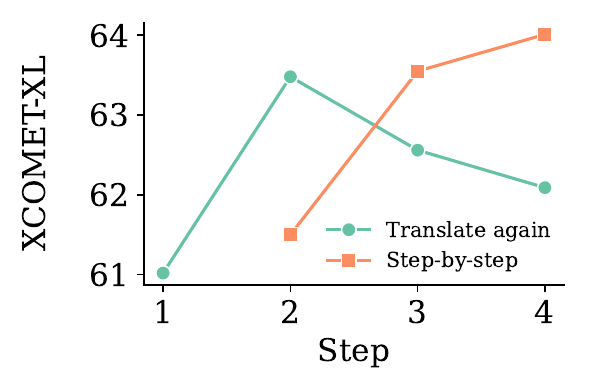}
        \caption{Gemini---Paragraph}
    \end{subfigure}

    \caption{{\color{myorange}\textit{Step-by-step}} vs. {\color{mygreen}\textit{Translate again}} results in MetricX and XCOMET-XL for GPT-4o-mini (top) and Gemini-2.0-Flash (bottom), for segment and paragraph-level translation. For MetricX, lower scores indicate a higher translation quality. See Fig. \ref{fig:fig-1} for an illustration and Appendix \ref{app:full_prompts} for full prompts.}
    \label{fig:metricx-xcomet-xl-comparison}
    \end{figure*}

    \definecolor{myorange}{RGB}{249,132,93}
    \definecolor{mygreen}{RGB}{85,164,139}

\begin{figure*}[t]
    \centering

    \begin{subfigure}[b]{0.23\linewidth}
        \includegraphics[width=\linewidth]{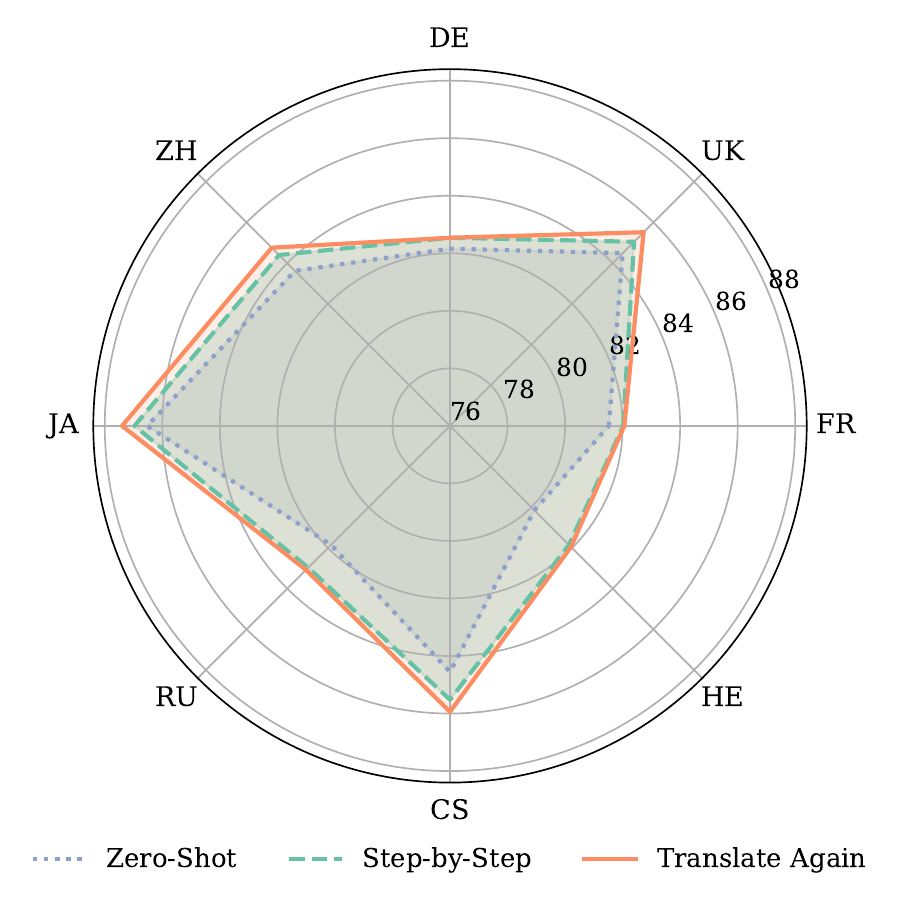}
        \caption{GPT---Seg: COMET}
    \end{subfigure}
    \hfill
    \begin{subfigure}[b]{0.23\linewidth}
        \includegraphics[width=\linewidth]{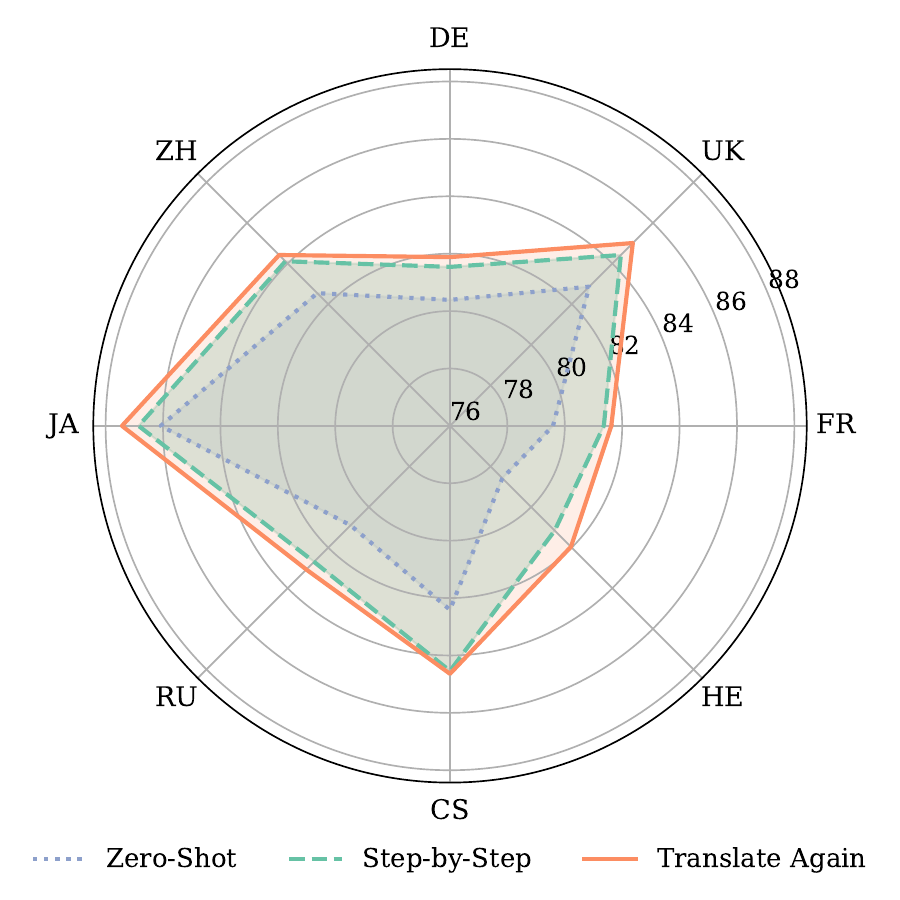}
        \caption{GPT---Par: COMET}
    \end{subfigure}
    \hfill
    \begin{subfigure}[b]{0.23\linewidth}
        \includegraphics[width=\linewidth]{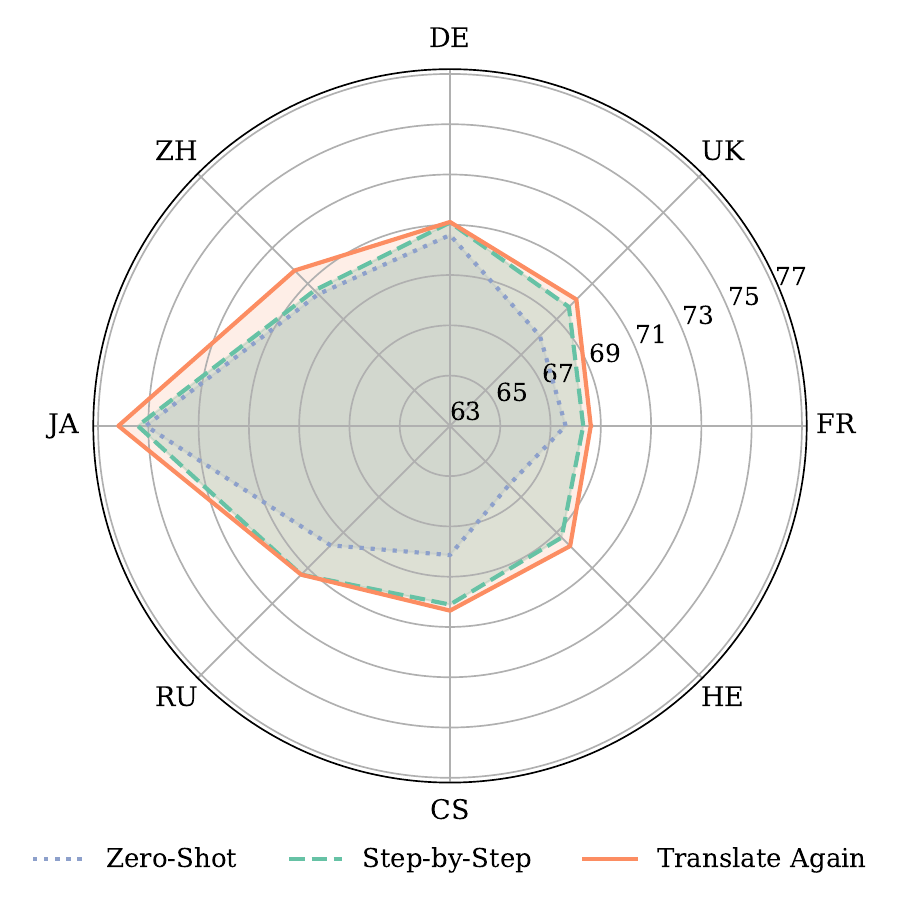}
        \caption{GPT---Seg: Kiwi-XL}
    \end{subfigure}
    \hfill
    \begin{subfigure}[b]{0.23\linewidth}
        \includegraphics[width=\linewidth]{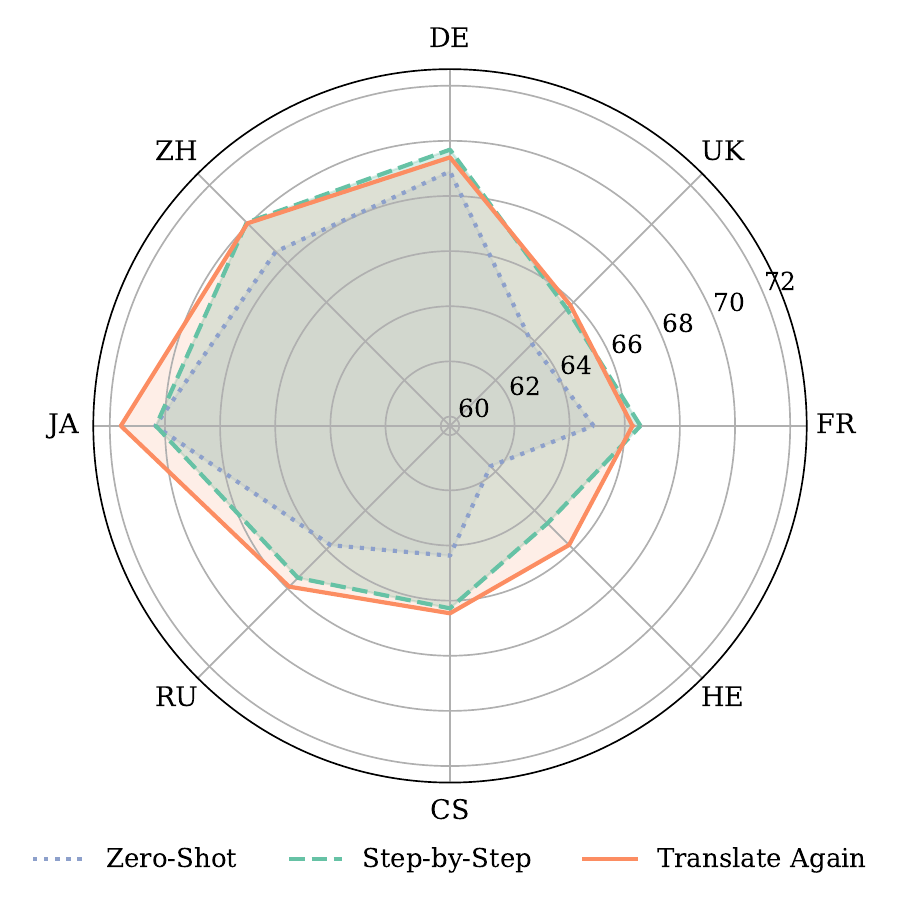}
        \caption{GPT---Par: Kiwi-XL}
    \end{subfigure}

    \vspace{1em}

    \begin{subfigure}[b]{0.23\linewidth}
        \includegraphics[width=\linewidth]{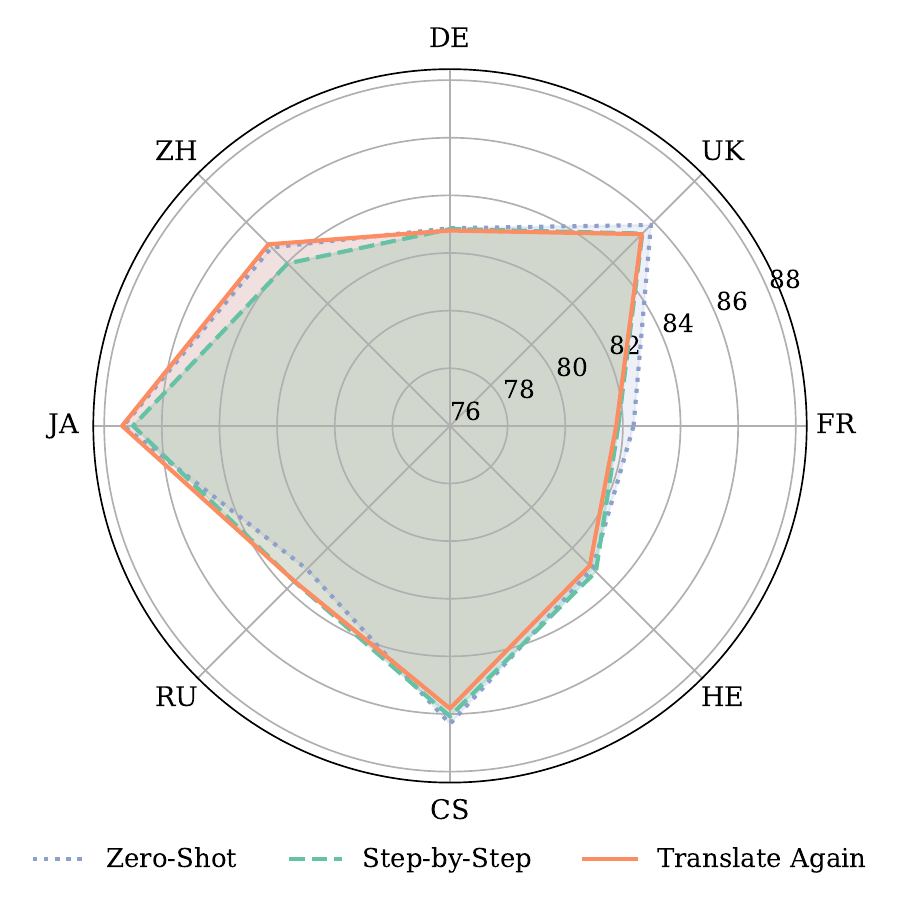}
        \caption{Gemini---Seg: COMET}
    \end{subfigure}
    \hfill
    \begin{subfigure}[b]{0.23\linewidth}
        \includegraphics[width=\linewidth]{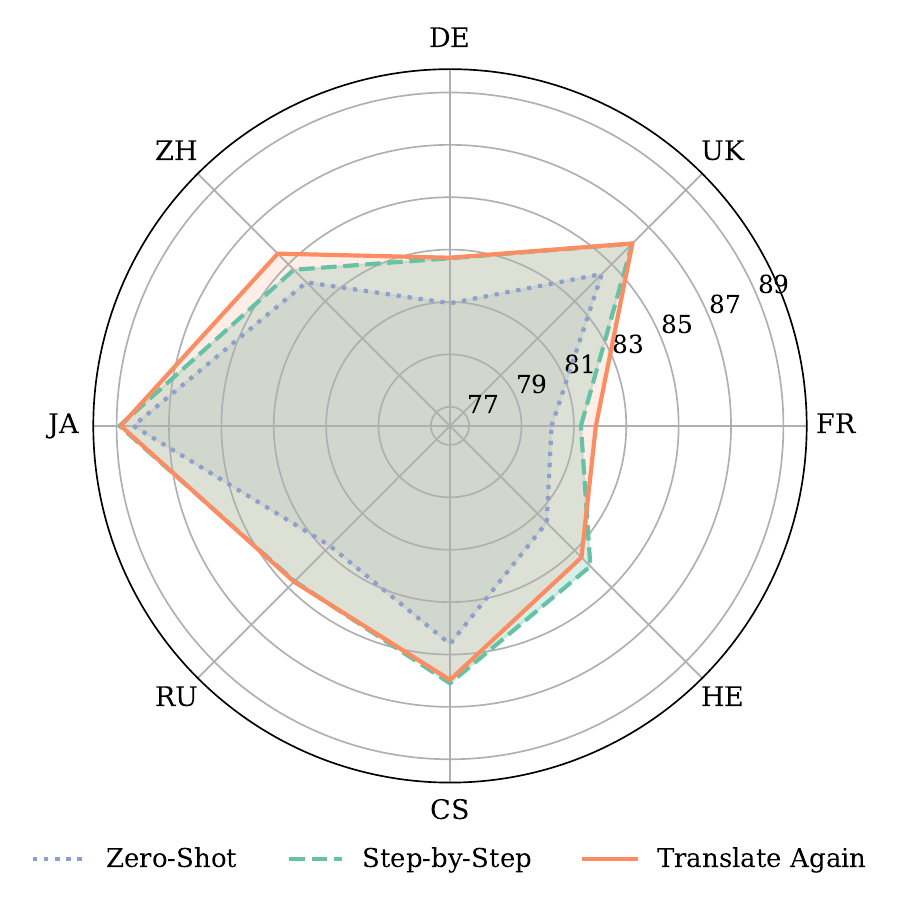}
        \caption{Gemini---Par: COMET}
    \end{subfigure}
    \hfill
    \begin{subfigure}[b]{0.23\linewidth}
        \includegraphics[width=\linewidth]{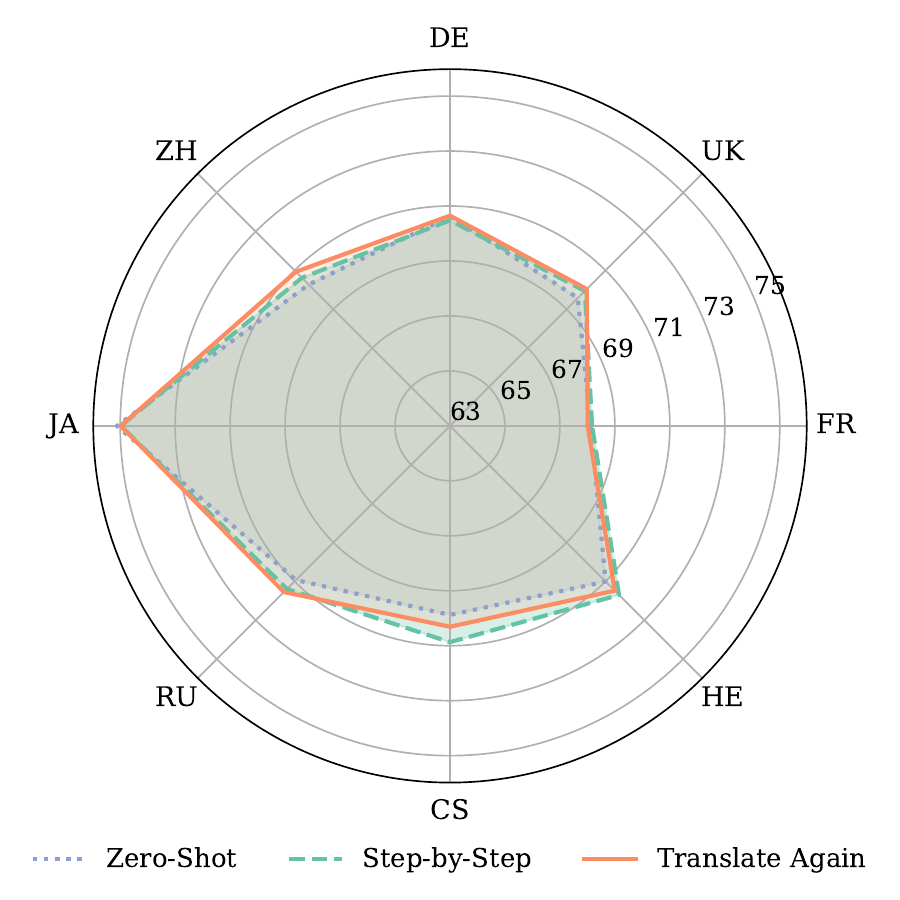}
        \caption{Gemini---Seg: Kiwi-XL}
    \end{subfigure}
    \hfill
    \begin{subfigure}[b]{0.23\linewidth}
        \includegraphics[width=\linewidth]{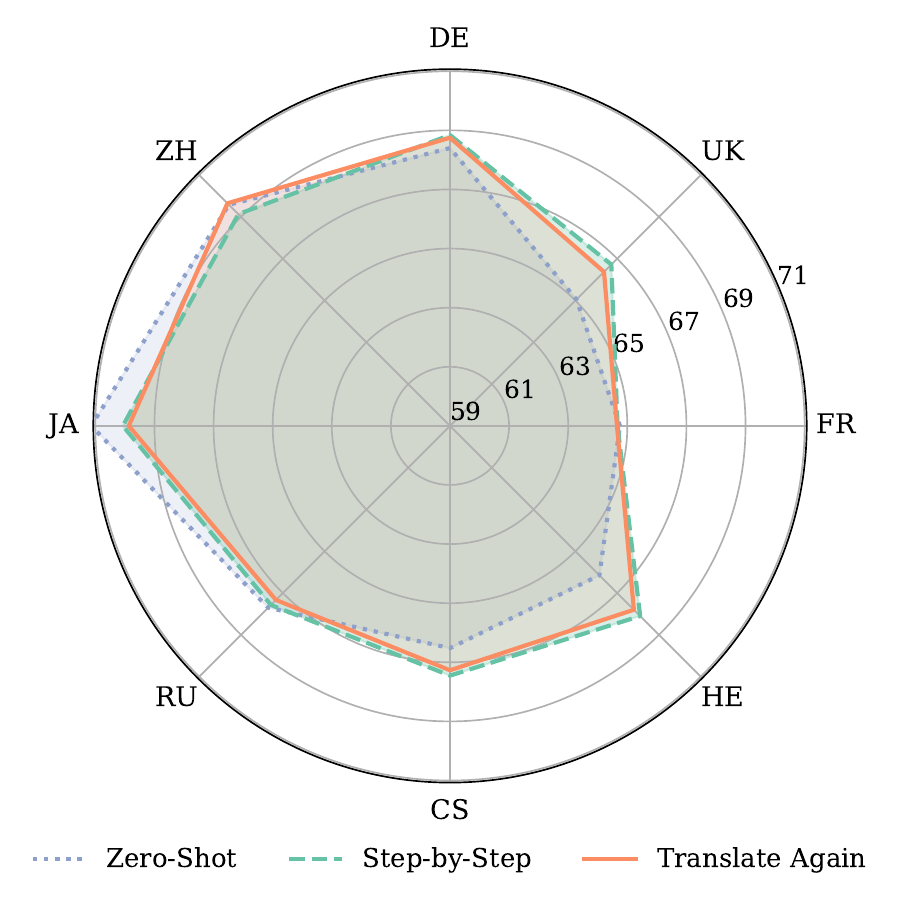}
        \caption{Gemini---Par: Kiwi-XL}
    \end{subfigure}

    \vspace{1em}

    \caption{COMET-22 and CometKiwi-XL results per-language for {\color{myorange}\textit{Step-by-step}} ({\color{myorange}Step 3}), {\color{mygreen}\textit{Translate again}} ({\color{mygreen}Step 2}), and Zero-Shot  ({\color{mygreen}Step 1}) prompts, for GPT-4o-mini (top) and Gemini-2.0-Flash (bottom), for segment and document-level translation.}
    \label{fig:radar-model-all}
\end{figure*}

\definecolor{lightmint}{RGB}{255, 255, 255}
\definecolor{forestgreen}{RGB}{34, 139, 34}
\begin{table*}[htbp]
\centering
\scriptsize
\begin{tabular}{llcccccccccc}
\toprule
Model & Setup & Step & \texttt{en\textrightarrow cs} & \texttt{en\textrightarrow de} & \texttt{en\textrightarrow fr} & \texttt{en\textrightarrow he} & \texttt{en\textrightarrow ja} & \texttt{en\textrightarrow ru} & \texttt{en\textrightarrow uk} & \texttt{en\textrightarrow zh} & Avg. \\
\midrule
\multirow{8}{*}{GPT-4o-mini} & \multirow{4}{*}{Step-by-step} & 1 & \multicolumn{1}{c}{--} & \multicolumn{1}{c}{--} & \multicolumn{1}{c}{--} & \multicolumn{1}{c}{--} & \multicolumn{1}{c}{--} & \multicolumn{1}{c}{--} & \multicolumn{1}{c}{--} & \multicolumn{1}{c}{--} & \multicolumn{1}{c}{--} \\
 &  & 2 & \cellcolor{forestgreen!62!lightmint}84.82 & \cellcolor{forestgreen!23!lightmint}82.19 & \cellcolor{forestgreen!10!lightmint}81.28 & \cellcolor{forestgreen!1!lightmint}80.72 & \cellcolor{forestgreen!84!lightmint}86.33 & \cellcolor{forestgreen!21!lightmint}82.06 & \cellcolor{forestgreen!55!lightmint}84.35 & \cellcolor{forestgreen!47!lightmint}83.83 & \cellcolor{forestgreen!38!lightmint}83.20 \\
 &  & 3 & \cellcolor{forestgreen!72!lightmint}85.52 & \cellcolor{forestgreen!28!lightmint}82.54 & \cellcolor{forestgreen!20!lightmint}82.02 & \cellcolor{forestgreen!18!lightmint}81.85 & \cellcolor{forestgreen!93!lightmint}86.97 & \cellcolor{forestgreen!34!lightmint}82.97 & \cellcolor{forestgreen!65!lightmint}85.05 & \cellcolor{forestgreen!55!lightmint}84.40 & \cellcolor{forestgreen!48!lightmint}83.91 \\
 &  & 4 & \cellcolor{forestgreen!76!lightmint}85.77 & \cellcolor{forestgreen!31!lightmint}82.74 & \cellcolor{forestgreen!23!lightmint}82.17 & \cellcolor{forestgreen!23!lightmint}82.16 & \cellcolor{forestgreen!95!lightmint}87.06 & \cellcolor{forestgreen!35!lightmint}83.01 & \cellcolor{forestgreen!69!lightmint}85.29 & \cellcolor{forestgreen!57!lightmint}84.48 & \cellcolor{forestgreen!51!lightmint}84.09 \\
\bottomrule
\end{tabular}
\caption{Full COMET-22 results for segment-level translation with GPT-4o-mini and Gemini-2.0-Flash, across 8 language pairs. Step-by-step Step 1 results are not shown since the model does not generate a translation at this step. A darker green shade indicates a better score.}
\label{tab:seg_comet}
\end{table*}

\definecolor{lightmint}{RGB}{255, 255, 255}
\definecolor{forestgreen}{RGB}{34, 139, 34}
\begin{table*}[htbp]
\centering
\scriptsize
\begin{tabular}{llcccccccccc}
\toprule
Model & Setup & Step & \texttt{en\textrightarrow cs} & \texttt{en\textrightarrow de} & \texttt{en\textrightarrow fr} & \texttt{en\textrightarrow he} & \texttt{en\textrightarrow ja} & \texttt{en\textrightarrow ru} & \texttt{en\textrightarrow uk} & \texttt{en\textrightarrow zh} & Avg. \\
\midrule
\multirow{8}{*}{GPT-4o-mini} & \multirow{4}{*}{Step-by-step} & 1 & \multicolumn{1}{c}{--} & \multicolumn{1}{c}{--} & \multicolumn{1}{c}{--} & \multicolumn{1}{c}{--} & \multicolumn{1}{c}{--} & \multicolumn{1}{c}{--} & \multicolumn{1}{c}{--} & \multicolumn{1}{c}{--} & \multicolumn{1}{c}{--} \\
 &  & 2 & \cellcolor{forestgreen!16!lightmint}68.76 & \cellcolor{forestgreen!35!lightmint}70.51 & \cellcolor{forestgreen!2!lightmint}67.44 & \cellcolor{forestgreen!0!lightmint}67.12 & \cellcolor{forestgreen!77!lightmint}74.29 & \cellcolor{forestgreen!33!lightmint}70.27 & \cellcolor{forestgreen!13!lightmint}68.46 & \cellcolor{forestgreen!27!lightmint}69.78 & \cellcolor{forestgreen!25!lightmint}69.58 \\
 &  & 3 & \cellcolor{forestgreen!31!lightmint}70.11 & \cellcolor{forestgreen!42!lightmint}71.09 & \cellcolor{forestgreen!11!lightmint}68.30 & \cellcolor{forestgreen!22!lightmint}69.28 & \cellcolor{forestgreen!89!lightmint}75.42 & \cellcolor{forestgreen!45!lightmint}71.37 & \cellcolor{forestgreen!26!lightmint}69.69 & \cellcolor{forestgreen!37!lightmint}70.62 & \cellcolor{forestgreen!38!lightmint}70.74 \\
 &  & 4 & \cellcolor{forestgreen!35!lightmint}70.46 & \cellcolor{forestgreen!43!lightmint}71.17 & \cellcolor{forestgreen!12!lightmint}68.41 & \cellcolor{forestgreen!28!lightmint}69.81 & \cellcolor{forestgreen!89!lightmint}75.40 & \cellcolor{forestgreen!47!lightmint}71.55 & \cellcolor{forestgreen!27!lightmint}69.71 & \cellcolor{forestgreen!39!lightmint}70.87 & \cellcolor{forestgreen!40!lightmint}70.92 \\
\bottomrule
\end{tabular}
\caption{Full CometKiwi-XL results for segment-level translation with GPT-4o-mini and Gemini-2.0-Flash, across 8 language pairs. Step-by-step Step 1 results are not shown since the model does not generate a translation at this step. A darker green shade indicates a better score.}
\label{tab:seg_kiwi-xl}
\end{table*}

\definecolor{lightmint}{RGB}{255, 255, 255}
\definecolor{forestgreen}{RGB}{34, 139, 34}
\begin{table*}[htbp]
\centering
\scriptsize
\begin{tabular}{llcccccccccc}
\toprule
Model & Setup & Step & \texttt{en\textrightarrow cs} & \texttt{en\textrightarrow de} & \texttt{en\textrightarrow fr} & \texttt{en\textrightarrow he} & \texttt{en\textrightarrow ja} & \texttt{en\textrightarrow ru} & \texttt{en\textrightarrow uk} & \texttt{en\textrightarrow zh} & Avg. \\
\midrule
\multirow{8}{*}{GPT-4o-mini} & \multirow{4}{*}{Step-by-step} & 1 & \multicolumn{1}{c}{--} & \multicolumn{1}{c}{--} & \multicolumn{1}{c}{--} & \multicolumn{1}{c}{--} & \multicolumn{1}{c}{--} & \multicolumn{1}{c}{--} & \multicolumn{1}{c}{--} & \multicolumn{1}{c}{--} & \multicolumn{1}{c}{--} \\
 &  & 2 & \cellcolor{forestgreen!40!lightmint}2.97 & \cellcolor{forestgreen!93!lightmint}1.45 & \cellcolor{forestgreen!46!lightmint}2.80 & \cellcolor{forestgreen!0!lightmint}4.66 & \cellcolor{forestgreen!45!lightmint}2.82 & \cellcolor{forestgreen!23!lightmint}3.46 & \cellcolor{forestgreen!22!lightmint}3.49 & \cellcolor{forestgreen!30!lightmint}3.25 & \cellcolor{forestgreen!35!lightmint}3.11 \\
 &  & 3 & \cellcolor{forestgreen!50!lightmint}2.68 & \cellcolor{forestgreen!96!lightmint}1.35 & \cellcolor{forestgreen!59!lightmint}2.42 & \cellcolor{forestgreen!0!lightmint}4.12 & \cellcolor{forestgreen!54!lightmint}2.57 & \cellcolor{forestgreen!34!lightmint}3.14 & \cellcolor{forestgreen!36!lightmint}3.08 & \cellcolor{forestgreen!38!lightmint}3.02 & \cellcolor{forestgreen!46!lightmint}2.80 \\
 &  & 4 & \cellcolor{forestgreen!52!lightmint}2.64 & \cellcolor{forestgreen!97!lightmint}1.33 & \cellcolor{forestgreen!61!lightmint}2.38 & \cellcolor{forestgreen!6!lightmint}3.94 & \cellcolor{forestgreen!55!lightmint}2.55 & \cellcolor{forestgreen!36!lightmint}3.10 & \cellcolor{forestgreen!38!lightmint}3.04 & \cellcolor{forestgreen!38!lightmint}3.02 & \cellcolor{forestgreen!48!lightmint}2.75 \\
\bottomrule
\end{tabular}
\caption{Full MetricX results for segment-level translation with GPT-4o-mini and Gemini-2.0-Flash, across 8 language pairs. Step-by-step Step 1 results are not shown since the model does not generate a translation at this step. A lower score and a darker green shade indicates better translation quality.}
\label{tab:seg_metricx}
\end{table*}

\definecolor{lightmint}{RGB}{255, 255, 255}
\definecolor{forestgreen}{RGB}{34, 139, 34}
\begin{table*}[htbp]
\centering
\scriptsize
\begin{tabular}{llcccccccccc}
\toprule
Model & Setup & Step & \texttt{en\textrightarrow cs} & \texttt{en\textrightarrow de} & \texttt{en\textrightarrow fr} & \texttt{en\textrightarrow he} & \texttt{en\textrightarrow ja} & \texttt{en\textrightarrow ru} & \texttt{en\textrightarrow uk} & \texttt{en\textrightarrow zh} & Avg. \\
\midrule
\multirow{8}{*}{GPT-4o-mini} & \multirow{4}{*}{Step-by-step} & 1 & \multicolumn{1}{c}{--} & \multicolumn{1}{c}{--} & \multicolumn{1}{c}{--} & \multicolumn{1}{c}{--} & \multicolumn{1}{c}{--} & \multicolumn{1}{c}{--} & \multicolumn{1}{c}{--} & \multicolumn{1}{c}{--} & \multicolumn{1}{c}{--} \\
 &  & 2 & \cellcolor{forestgreen!31!lightmint}81.10 & \cellcolor{forestgreen!91!lightmint}90.40 & \cellcolor{forestgreen!34!lightmint}81.70 & \cellcolor{forestgreen!0!lightmint}73.60 & \cellcolor{forestgreen!7!lightmint}77.52 & \cellcolor{forestgreen!32!lightmint}81.25 & \cellcolor{forestgreen!21!lightmint}79.63 & \cellcolor{forestgreen!0!lightmint}75.37 & \cellcolor{forestgreen!24!lightmint}80.07 \\
 &  & 3 & \cellcolor{forestgreen!42!lightmint}82.86 & \cellcolor{forestgreen!95!lightmint}91.06 & \cellcolor{forestgreen!44!lightmint}83.24 & \cellcolor{forestgreen!1!lightmint}76.60 & \cellcolor{forestgreen!17!lightmint}78.96 & \cellcolor{forestgreen!42!lightmint}82.93 & \cellcolor{forestgreen!32!lightmint}81.37 & \cellcolor{forestgreen!5!lightmint}77.19 & \cellcolor{forestgreen!35!lightmint}81.78 \\
 &  & 4 & \cellcolor{forestgreen!45!lightmint}83.28 & \cellcolor{forestgreen!96!lightmint}91.20 & \cellcolor{forestgreen!45!lightmint}83.35 & \cellcolor{forestgreen!6!lightmint}77.33 & \cellcolor{forestgreen!17!lightmint}79.06 & \cellcolor{forestgreen!43!lightmint}82.94 & \cellcolor{forestgreen!35!lightmint}81.75 & \cellcolor{forestgreen!6!lightmint}77.39 & \cellcolor{forestgreen!37!lightmint}82.04 \\
\bottomrule
\end{tabular}
\caption{Full XCOMET-XL results for segment-level translation with GPT-4o-mini and Gemini-2.0-Flash, across 8 language pairs. Step-by-step Step 1 results are not shown since the model does not generate a translation at this step. A darker green shade indicates a better score.}
\label{tab:seg_xcomet-xl}
\end{table*}

\definecolor{lightmint}{RGB}{255, 255, 255}
\definecolor{forestgreen}{RGB}{34, 139, 34}
\begin{table*}[htbp]
\centering
\scriptsize
\begin{tabular}{llcccccccccc}
\toprule
Model & Setup & Step & \texttt{en\textrightarrow cs} & \texttt{en\textrightarrow de} & \texttt{en\textrightarrow fr} & \texttt{en\textrightarrow he} & \texttt{en\textrightarrow ja} & \texttt{en\textrightarrow ru} & \texttt{en\textrightarrow uk} & \texttt{en\textrightarrow zh} & Avg. \\
\midrule
\multirow{8}{*}{GPT-4o-mini} & \multirow{4}{*}{Step-by-step} & 1 & \multicolumn{1}{c}{--} & \multicolumn{1}{c}{--} & \multicolumn{1}{c}{--} & \multicolumn{1}{c}{--} & \multicolumn{1}{c}{--} & \multicolumn{1}{c}{--} & \multicolumn{1}{c}{--} & \multicolumn{1}{c}{--} & \multicolumn{1}{c}{--} \\
 &  & 2 & \cellcolor{forestgreen!23!lightmint}83.14 & \cellcolor{forestgreen!0!lightmint}80.41 & \cellcolor{forestgreen!0!lightmint}80.49 & \cellcolor{forestgreen!0!lightmint}78.15 & \cellcolor{forestgreen!60!lightmint}85.92 & \cellcolor{forestgreen!0!lightmint}81.32 & \cellcolor{forestgreen!23!lightmint}83.12 & \cellcolor{forestgreen!12!lightmint}82.31 & \cellcolor{forestgreen!6!lightmint}81.86 \\
 &  & 3 & \cellcolor{forestgreen!42!lightmint}84.54 & \cellcolor{forestgreen!2!lightmint}81.54 & \cellcolor{forestgreen!0!lightmint}81.36 & \cellcolor{forestgreen!0!lightmint}81.16 & \cellcolor{forestgreen!72!lightmint}86.85 & \cellcolor{forestgreen!17!lightmint}82.70 & \cellcolor{forestgreen!40!lightmint}84.42 & \cellcolor{forestgreen!36!lightmint}84.11 & \cellcolor{forestgreen!26!lightmint}83.33 \\
 &  & 4 & \cellcolor{forestgreen!44!lightmint}84.70 & \cellcolor{forestgreen!3!lightmint}81.65 & \cellcolor{forestgreen!2!lightmint}81.52 & \cellcolor{forestgreen!3!lightmint}81.61 & \cellcolor{forestgreen!68!lightmint}86.56 & \cellcolor{forestgreen!23!lightmint}83.13 & \cellcolor{forestgreen!47!lightmint}84.97 & \cellcolor{forestgreen!35!lightmint}84.02 & \cellcolor{forestgreen!28!lightmint}83.52 \\
\bottomrule
\end{tabular}
\caption{Full COMET-22 results for paragraph-level translation with GPT-4o-mini and Gemini-2.0-Flash, across 8 language pairs. Step-by-step Step 1 results are not shown since the model does not generate a translation at this step. A darker green shade indicates a better score.}
\label{tab:doc_comet}
\end{table*}

\definecolor{lightmint}{RGB}{255, 255, 255}
\definecolor{forestgreen}{RGB}{34, 139, 34}
\begin{table*}[htbp]
\centering
\scriptsize
\begin{tabular}{llcccccccccc}
\toprule
Model & Setup & Step & \texttt{en\textrightarrow cs} & \texttt{en\textrightarrow de} & \texttt{en\textrightarrow fr} & \texttt{en\textrightarrow he} & \texttt{en\textrightarrow ja} & \texttt{en\textrightarrow ru} & \texttt{en\textrightarrow uk} & \texttt{en\textrightarrow zh} & Avg. \\
\midrule
\multirow{8}{*}{GPT-4o-mini} & \multirow{4}{*}{Step-by-step} & 1 & \multicolumn{1}{c}{--} & \multicolumn{1}{c}{--} & \multicolumn{1}{c}{--} & \multicolumn{1}{c}{--} & \multicolumn{1}{c}{--} & \multicolumn{1}{c}{--} & \multicolumn{1}{c}{--} & \multicolumn{1}{c}{--} & \multicolumn{1}{c}{--} \\
 &  & 2 & \cellcolor{forestgreen!10!lightmint}64.50 & \cellcolor{forestgreen!62!lightmint}68.71 & \cellcolor{forestgreen!26!lightmint}65.81 & \cellcolor{forestgreen!0!lightmint}60.03 & \cellcolor{forestgreen!70!lightmint}69.34 & \cellcolor{forestgreen!25!lightmint}65.69 & \cellcolor{forestgreen!3!lightmint}63.90 & \cellcolor{forestgreen!52!lightmint}67.87 & \cellcolor{forestgreen!25!lightmint}65.73 \\
 &  & 3 & \cellcolor{forestgreen!32!lightmint}66.28 & \cellcolor{forestgreen!74!lightmint}69.68 & \cellcolor{forestgreen!36!lightmint}66.57 & \cellcolor{forestgreen!12!lightmint}64.66 & \cellcolor{forestgreen!82!lightmint}70.33 & \cellcolor{forestgreen!47!lightmint}67.46 & \cellcolor{forestgreen!25!lightmint}65.66 & \cellcolor{forestgreen!79!lightmint}70.09 & \cellcolor{forestgreen!48!lightmint}67.59 \\
 &  & 4 & \cellcolor{forestgreen!35!lightmint}66.47 & \cellcolor{forestgreen!75!lightmint}69.77 & \cellcolor{forestgreen!36!lightmint}66.57 & \cellcolor{forestgreen!24!lightmint}65.60 & \cellcolor{forestgreen!77!lightmint}69.91 & \cellcolor{forestgreen!51!lightmint}67.83 & \cellcolor{forestgreen!32!lightmint}66.22 & \cellcolor{forestgreen!77!lightmint}69.95 & \cellcolor{forestgreen!51!lightmint}67.79 \\
\bottomrule
\end{tabular}
\caption{Full CometKiwi-XL results for paragraph-level translation with GPT-4o-mini and Gemini-2.0-Flash, across 8 language pairs. Step-by-step Step 1 results are not shown since the model does not generate a translation at this step. A darker green shade indicates a better score.}
\label{tab:doc_kiwi-xl}
\end{table*}

\definecolor{lightmint}{RGB}{255, 255, 255}
\definecolor{forestgreen}{RGB}{34, 139, 34}
\begin{table*}[htbp]
\centering
\scriptsize
\begin{tabular}{llcccccccccc}
\toprule
Model & Setup & Step & \texttt{en\textrightarrow cs} & \texttt{en\textrightarrow de} & \texttt{en\textrightarrow fr} & \texttt{en\textrightarrow he} & \texttt{en\textrightarrow ja} & \texttt{en\textrightarrow ru} & \texttt{en\textrightarrow uk} & \texttt{en\textrightarrow zh} & Avg. \\
\midrule
\multirow{8}{*}{GPT-4o-mini} & \multirow{4}{*}{Step-by-step} & 1 & \multicolumn{1}{c}{--} & \multicolumn{1}{c}{--} & \multicolumn{1}{c}{--} & \multicolumn{1}{c}{--} & \multicolumn{1}{c}{--} & \multicolumn{1}{c}{--} & \multicolumn{1}{c}{--} & \multicolumn{1}{c}{--} & \multicolumn{1}{c}{--} \\
 &  & 2 & \cellcolor{forestgreen!18!lightmint}8.16 & \cellcolor{forestgreen!89!lightmint}3.10 & \cellcolor{forestgreen!40!lightmint}6.58 & \cellcolor{forestgreen!0!lightmint}11.76 & \cellcolor{forestgreen!32!lightmint}7.19 & \cellcolor{forestgreen!14!lightmint}8.41 & \cellcolor{forestgreen!17!lightmint}8.21 & \cellcolor{forestgreen!26!lightmint}7.57 & \cellcolor{forestgreen!25!lightmint}7.62 \\
 &  & 3 & \cellcolor{forestgreen!32!lightmint}7.18 & \cellcolor{forestgreen!93!lightmint}2.81 & \cellcolor{forestgreen!50!lightmint}5.85 & \cellcolor{forestgreen!0!lightmint}10.20 & \cellcolor{forestgreen!40!lightmint}6.61 & \cellcolor{forestgreen!24!lightmint}7.75 & \cellcolor{forestgreen!27!lightmint}7.51 & \cellcolor{forestgreen!34!lightmint}6.99 & \cellcolor{forestgreen!36!lightmint}6.86 \\
 &  & 4 & \cellcolor{forestgreen!34!lightmint}7.01 & \cellcolor{forestgreen!94!lightmint}2.77 & \cellcolor{forestgreen!51!lightmint}5.79 & \cellcolor{forestgreen!0!lightmint}9.67 & \cellcolor{forestgreen!40!lightmint}6.58 & \cellcolor{forestgreen!27!lightmint}7.52 & \cellcolor{forestgreen!30!lightmint}7.27 & \cellcolor{forestgreen!35!lightmint}6.93 & \cellcolor{forestgreen!39!lightmint}6.69 \\
\bottomrule
\end{tabular}
\caption{Full MetricX results for paragraph-level translation with GPT-4o-mini and Gemini-2.0-Flash, across 8 language pairs. Step-by-step Step 1 results are not shown since the model does not generate a translation at this step. A lower score and a darker green shade indicates better translation quality.}
\label{tab:doc_metricx}
\end{table*}

\definecolor{lightmint}{RGB}{255, 255, 255}
\definecolor{forestgreen}{RGB}{34, 139, 34}
\begin{table*}[htbp]
\centering
\scriptsize
\begin{tabular}{llcccccccccc}
\toprule
Model & Setup & Step & \texttt{en\textrightarrow cs} & \texttt{en\textrightarrow de} & \texttt{en\textrightarrow fr} & \texttt{en\textrightarrow he} & \texttt{en\textrightarrow ja} & \texttt{en\textrightarrow ru} & \texttt{en\textrightarrow uk} & \texttt{en\textrightarrow zh} & Avg. \\
\midrule
\multirow{8}{*}{GPT-4o-mini} & \multirow{4}{*}{Step-by-step} & 1 & \multicolumn{1}{c}{--} & \multicolumn{1}{c}{--} & \multicolumn{1}{c}{--} & \multicolumn{1}{c}{--} & \multicolumn{1}{c}{--} & \multicolumn{1}{c}{--} & \multicolumn{1}{c}{--} & \multicolumn{1}{c}{--} & \multicolumn{1}{c}{--} \\
 &  & 2 & \cellcolor{forestgreen!20!lightmint}55.70 & \cellcolor{forestgreen!89!lightmint}73.35 & \cellcolor{forestgreen!24!lightmint}56.62 & \cellcolor{forestgreen!0!lightmint}43.45 & \cellcolor{forestgreen!9!lightmint}52.92 & \cellcolor{forestgreen!29!lightmint}58.07 & \cellcolor{forestgreen!22!lightmint}56.14 & \cellcolor{forestgreen!12!lightmint}53.49 & \cellcolor{forestgreen!22!lightmint}56.22 \\
 &  & 3 & \cellcolor{forestgreen!34!lightmint}59.16 & \cellcolor{forestgreen!95!lightmint}75.02 & \cellcolor{forestgreen!32!lightmint}58.65 & \cellcolor{forestgreen!0!lightmint}49.22 & \cellcolor{forestgreen!18!lightmint}55.03 & \cellcolor{forestgreen!44!lightmint}61.86 & \cellcolor{forestgreen!32!lightmint}58.75 & \cellcolor{forestgreen!26!lightmint}57.19 & \cellcolor{forestgreen!34!lightmint}59.36 \\
 &  & 4 & \cellcolor{forestgreen!36!lightmint}59.65 & \cellcolor{forestgreen!96!lightmint}75.19 & \cellcolor{forestgreen!31!lightmint}58.40 & \cellcolor{forestgreen!2!lightmint}50.98 & \cellcolor{forestgreen!19!lightmint}55.34 & \cellcolor{forestgreen!46!lightmint}62.44 & \cellcolor{forestgreen!37!lightmint}60.02 & \cellcolor{forestgreen!26!lightmint}57.28 & \cellcolor{forestgreen!37!lightmint}59.91 \\
\bottomrule
\end{tabular}
\caption{Full XCOMET-XL results for paragraph-level translation with GPT-4o-mini and Gemini-2.0-Flash, across 8 language pairs. Step-by-step Step 1 results are not shown since the model does not generate a translation at this step. A darker green shade indicates a better score.}
\label{tab:doc_xcomet-xl}
\end{table*}

\begin{figure*}[t]
\centering
\begin{tcolorbox}
\label{gpt_prompt}
\textbf{System:}
You are a helpful assistant.
\vspace{1em}

\textbf{User:}
You will be asked to translate a piece of text from [source language] into [target language] following stages of the translation process. Here is the context in which the text appears:

\vspace{1em}
Context: [source text]
\vspace{1em}

To start, let’s do some pre-drafting research on the above context.
\vspace{1em}

Research: During this phase, thorough research is essential to address components of the context text that pose translation challenges. The goal is to establish a comprehensive translation plan that covers the following categories:
\vspace{1em}

Idiomatic Expressions:
\vspace{1em}

\hspace{2em} - Identify idiomatic expressions that cannot be directly translated word-for-word into [target language].
\end{tcolorbox}
\caption{Prompt template used in the research (decomposition) stage of \textit{step-by-step} translation.}
\label{app:step-by-step-1}
\end{figure*}

\begin{figure*}[t]
\centering
\begin{tcolorbox}
\label{gpt_prompt}
\textbf{System:}
You are a helpful assistant.
\vspace{1em}

\textbf{User:}
Now, let’s move on to the drafting stage.
\vspace{1em}

Draft Translation:
\vspace{1em}

In this phase, your primary objective is to create a draft translation that accurately conveys the meaning of the source text presented below. At this stage, it is crucial to focus on adequacy, ensuring that your translation closely adheres to the source text. Your response should conclude with the draft translation. If context is missing, generate a general translation that is adaptable to various contexts. Avoid adding any additional information not present in the source text. All elements of the source text should be present in the translation.

Provide your single best translation of the following text, guided by the pre-drafting analysis, without adding anything further:
\vspace{1em}

English: [source text]
\end{tcolorbox}
\caption{Prompt used in the drafting (translation) stage of \textit{step-by-step} translation.}
\label{app:step-by-step-2}
\end{figure*}

\begin{figure*}[t]
\centering
\begin{tcolorbox}
\textbf{System:}
You are a helpful assistant.
\vspace{1em}

\textbf{User:}
Now let’s move to the next stage.
\vspace{1em}

Post-editing with local refinement: In this stage, the primary aim is to refine the draft translation by making micro-level improvements that improve the draft’s fluency.
\vspace{1em}

Provide only one refined translation and do not output anything else after that.
\end{tcolorbox}
\caption{Prompt used in the post-editing (refinement) stage of \textit{step-by-step} translation.}
\label{app:step-by-step-3}
\end{figure*}

\begin{figure*}[t]
\centering
\begin{tcolorbox}
\textbf{System:} You are a helpful assistant.
\vspace{1em}

\textbf{User:} You are tasked with proofreading a translation that has been revised for improved fluency. The refined translation has been generated by editing the draft translation.
\vspace{1em}

Proofreading and Final Editing: The goal is to provide a polished final translation of the source text. For your reference, below are the source text, the draft, and refined translations.
\vspace{1em}

\textbf{Source Text:} [source text]

\textbf{Draft Translation:} [Step 2 output]

\textbf{Refined Translation:} [Step 3 output]
\vspace{1em}

Please proofread the refined text for grammar, spelling, punctuation, terminology, and overall fluency. Ensure the translation accurately reflects the original meaning and style. Provide only the final, polished translation on the first line.
\end{tcolorbox}
\caption{Prompt used in the proofreading stage of \textit{step-by-step} translation.}
\label{app:step-by-step-4}
\end{figure*}

\begin{figure*}[t]
\centering
\begin{tcolorbox}
\textbf{System:}
You are a helpful assistant.
\vspace{1em}

\textbf{User:}
Please translate the following text from [source language] to [target language]. Provide only one translation and do not output anything else after that.
\vspace{1em}

English: [source text]
\end{tcolorbox}
\caption{Prompt used in the translation stage of \textit{translate again} prompting.}
\label{app:translate-again-1}
\end{figure*}

\begin{figure*}[t]
\centering
\begin{tcolorbox}
\textbf{System:}
You are a helpful assistant.
\vspace{1em}

\textbf{User:}
Please again translate the following text from [source language] to [target language] to make it better. Provide only one translation and do not output anything else after that.
\vspace{1em}

English: [source text]

\end{tcolorbox}
\caption{Prompt used in the refinement stage of \textit{translate again} prompting. In this prompt, the model is provided with all previous prompts and outputs as part of a multi-turn conversation.}
\label{app:translate-again-2}
\end{figure*}

\begin{figure*}[t]
\centering
\begin{tcolorbox}[width=\textwidth, colback=gray!5, colframe=black!40, boxrule=0.5pt, sharp corners=south]
\textbf{System:} You are a helpful assistant.

\vspace{1em}

\textbf{User:} Given the following English original text and the corresponding analysis:

\vspace{1em}
English Original Text: [source text]

\vspace{1em}
Analysis: [analysis]

\vspace{1em}
Please analyze the differences between the following two translations in \{tgt\_lang\}:

\vspace{1em}
Translation-1: [translation 1]\\
Translation-2: [translation 2]

\vspace{1em}
1. First, list the main differences between Translation-1 and Translation-2 in terms of wording, syntax, semantics, or style. Present the differences as a numbered list.

\vspace{1em}
2. For each difference, state whether it is explicitly or implicitly addressed in the Analysis. If yes, mention the corresponding part of the analysis.

\vspace{1em}
3. Count how many of the differences related to Translation-1 are reflected in the analysis, and how many related to Translation-2 are reflected.

\vspace{1em}
4. Output only the following two tags on the last line:

\texttt{<trans-1-cnt>number</trans-1-cnt>} and \texttt{<trans-2-cnt>number</trans-2-cnt>}

\end{tcolorbox}
\caption{Prompt used for \textit{LLM-as-a-Judge}.}
\label{app:llm-as-a-judge}
\end{figure*}

\end{document}